\documentclass[manuscript,screen]{acmart}
\AtBeginDocument{%
  \providecommand\BibTeX{{%
    \normalfont B\kern-0.5em{\scshape i\kern-0.25em b}\kern-0.8em\TeX}}}

\setcopyright{acmlicensed}
\acmJournal{CSUR}
\acmYear{2020} \acmVolume{1} \acmNumber{1} \acmArticle{1} \acmMonth{1} \acmPrice{15.00}\acmDOI{10.1145/3409383}

\acmConference[CSUR]{ACM Computing Surveys}{March, 2020}{New York, NY}
\acmBooktitle{ACM Computing Surveys}
\acmPrice{15.00}
\acmISBN{xxx-x-xxxx-xxxx-x/xx/xx}

\usepackage{amsmath}
\usepackage[none]{hyphenat}
\usepackage{epsfig,graphicx,subfigure}
\usepackage{multirow}
\usepackage{multicol}
\usepackage{amsmath}
\usepackage{amsfonts}
\usepackage{subfigure}
\usepackage{graphicx}
\usepackage{multirow}
\usepackage{url}
\usepackage{paralist}
\usepackage{mathrsfs, euscript}
\usepackage{color}
\usepackage{xcolor}
\usepackage{pifont}
\begin{document}
\def\Blue{\color{blue}}
\def\Purple{\color{purple}}

\def\A{{\bf A}}
\def\a{{\bf a}}
\def\B{{\bf B}}
\def\b{{\bf b}}
\def\C{{\bf C}}
\def\c{{\bf c}}
\def\D{{\bf D}}
\def\d{{\bf d}}
\def\E{{\bf E}}
\def\e{{\bf e}}
\def\f{{\bf f}}
\def\F{{\bf F}}
\def\K{{\bf K}}
\def\k{{\bf k}}
\def\L{{\bf L}}
\def\H{{\bf H}}
\def\h{{\bf h}}
\def\G{{\bf G}}
\def\g{{\bf g}}
\def\I{{\bf I}}
\def\J{{\bf J}}
\def\R{{\bf R}}
\def\X{{\bf X}}
\def\Y{{\bf Y}}
\def\OO{{\bf O}}
\def\oo{{\bf o}}
\def\P{{\bf P}}
\def\p{{\bf p}}
\def\Q{{\bf Q}}
\def\q{{\bf q}}
\def\r{{\bf r}}
\def\s{{\bf s}}
\def\S{{\bf S}}
\def\t{{\bf t}}
\def\T{{\bf T}}
\def\x{{\bf x}}
\def\y{{\bf y}}
\def\z{{\bf z}}
\def\Z{{\bf Z}}
\def\M{{\bf M}}
\def\m{{\bf m}}
\def\n{{\bf n}}
\def\U{{\bf U}}
\def\u{{\bf u}}
\def\V{{\bf V}}
\def\v{{\bf v}}
\def\W{{\bf W}}
\def\w{{\bf w}}
\def\0{{\bf 0}}
\def\1{{\bf 1}}

\def\AM{{\mathcal A}}
\def\EM{{\mathcal E}}
\def\FM{{\mathcal F}}
\def\TM{{\mathcal T}}
\def\UM{{\mathcal U}}
\def\XM{{\mathcal X}}
\def\YM{{\mathcal Y}}
\def\NM{{\mathcal N}}
\def\OM{{\mathcal O}}
\def\IM{{\mathcal I}}
\def\GM{{\mathcal G}}
\def\PM{{\mathcal P}}
\def\LM{{\mathcal L}}
\def\MM{{\mathcal M}}
\def\DM{{\mathcal D}}
\def\SM{{\mathcal S}}
\def\RB{{\mathbb R}}
\def\EB{{\mathbb E}}

\def\tx{\tilde{\bf x}}
\def\ty{\tilde{\bf y}}
\def\tz{\tilde{\bf z}}
\def\hd{\hat{d}}
\def\HD{\hat{\bf D}}
\def\hx{\hat{\bf x}}
\def\hR{\hat{R}}

\def\Ome{\mbox{\boldmath$\omega$\unboldmath}}
\def\Om{\mbox{\boldmath$\Omega$\unboldmath}}
\def\bet{\mbox{\boldmath$\beta$\unboldmath}}
\def\et{\mbox{\boldmath$\eta$\unboldmath}}
\def\ep{\mbox{\boldmath$\epsilon$\unboldmath}}
\def\ph{\mbox{\boldmath$\phi$\unboldmath}}
\def\Pii{\mbox{\boldmath$\Pi$\unboldmath}}
\def\pii{\mbox{\boldmath$\pi$\unboldmath}}
\def\Ph{\mbox{\boldmath$\Phi$\unboldmath}}
\def\Ps{\mbox{\boldmath$\Psi$\unboldmath}}
\def\tha{\mbox{\boldmath$\theta$\unboldmath}}
\def\Tha{\mbox{\boldmath$\Theta$\unboldmath}}
\def\muu{\mbox{\boldmath$\mu$\unboldmath}}
\def\Si{\mbox{\boldmath$\Sigma$\unboldmath}}
\def\si{\mbox{\boldmath$\sigma$\unboldmath}}
\def\Gam{\mbox{\boldmath$\Gamma$\unboldmath}}
\def\gamm{\mbox{\boldmath$\gamma$\unboldmath}}
\def\Lam{\mbox{\boldmath$\Lambda$\unboldmath}}
\def\De{\mbox{\boldmath$\Delta$\unboldmath}}
\def\vps{\mbox{\boldmath$\varepsilon$\unboldmath}}
\def\Up{\mbox{\boldmath$\Upsilon$\unboldmath}}
\def\xii{\mbox{\boldmath$\xi$\unboldmath}}
\def\Xii{\mbox{\boldmath$\Xi$\unboldmath}}
\def\Lap{\mbox{\boldmath$\LM$\unboldmath}}
\newcommand{\ti}[1]{\tilde{#1}}

\def\tr{\mathrm{tr}}
\def\etr{\mathrm{etr}}
\def\etal{{\em et al.\/}\,}
\newcommand{\indep}{{\;\bot\!\!\!\!\!\!\bot\;}}
\def\argmax{\mathop{\rm argmax}}
\def\argmin{\mathop{\rm argmin}}
\def\vec{\text{vec}}
\def\cov{\text{cov}}
\def\dg{\text{diag}}

\title{A Survey on Bayesian Deep Learning}

\author{Hao Wang}
\email{hoguewang@gmail.com}
\affiliation{%
  \institution{Massachusetts Institute of Technology}
  \country{USA}
}

\author{Dit-Yan Yeung}
\email{dyyeung\}@cse.ust.hk}
\affiliation{%
  \institution{Hong Kong University of Science and Technology}
  \streetaddress{P.O. Box 1212}
  \country{Hong Kong}
}

\begin{abstract}
A comprehensive artificial intelligence system needs to not only perceive the environment with different `senses' (e.g., seeing and hearing) but also infer the world's conditional (or even causal) relations and corresponding uncertainty. The past decade has seen major advances in many perception tasks such as visual object recognition and speech recognition using deep learning models. For higher-level inference, however, probabilistic graphical models with their Bayesian nature are still more powerful and flexible. In recent years, \emph{Bayesian deep learning} has emerged as a unified probabilistic framework to tightly integrate deep learning and Bayesian models\footnote{See a curated and updating list of papers related to Bayesian deep learning at \url{https://github.com/js05212/BayesianDeepLearning-Survey}.}. In this general framework, the perception of text or images using deep learning can boost the performance of higher-level inference and in turn, the feedback from the inference process is able to enhance the perception of text or images. This survey provides a comprehensive introduction to \emph{Bayesian deep learning} and reviews its recent applications on recommender systems, topic models, control, etc. Besides, we also discuss the relationship and differences between Bayesian deep learning and other related topics such as Bayesian treatment of neural networks. 
\end{abstract}




\maketitle

\section{Introduction}\label{sec:intro}
Over the past decade, deep learning has achieved significant success in many popular perception tasks including visual object recognition, text understanding, and speech recognition. These tasks correspond to artificial intelligence (AI) systems' ability to \emph{see}, \emph{read}, and \emph{hear}, respectively, and they are undoubtedly indispensable for AI to effectively perceive the environment. However, in order to build a practical and comprehensive AI system, simply being able to perceive is far from sufficient. It should, above all, possess the ability of \emph{thinking}. 
% Deep learning has achieved significant success in many perception tasks including \emph{seeing} (visual object recognition), \emph{reading} (text understanding), and \emph{hearing} (speech recognition). These are undoubtedly fundamental tasks for a functioning comprehensive artificial intelligence (AI) system. However, in order to build a real AI system, simply being able to see, read, and hear is far from enough. It should, above all, possess the ability of \emph{thinking}.

A typical example is medical diagnosis, which goes far beyond simple perception: besides \emph{seeing} visible symptoms (or medical images from CT) and \emph{hearing} descriptions from patients, a doctor also has to look for relations among all the symptoms and preferably infer their corresponding etiology. Only after that can the doctor provide medical advice for the patients. In this example, although the abilities of \emph{seeing} and \emph{hearing} allow the doctor to acquire information from the patients, it is the \emph{thinking} part that defines a doctor. Specifically, the ability of \emph{thinking} here could involve identifying conditional dependencies, causal inference, logic deduction, and dealing with uncertainty, which are apparently beyond the capability of conventional deep learning methods. Fortunately, another machine learning paradigm, probabilistic graphical models (PGM), excels at probabilistic or causal inference and at dealing with uncertainty. The problem is that PGM is not as good as deep learning models at perception tasks, which usually involve large-scale and high-dimensional signals (e.g., images and videos). To address this problem, it is therefore a natural choice to unify deep learning and PGM within a principled probabilistic framework, which we call \emph{Bayesian deep learning} (BDL) in this paper.

In the example above, the \emph{perception task} involves perceiving the patient's symptoms (e.g., by \emph{seeing} medical images), while the \emph{inference task} involves handling conditional dependencies, causal inference, logic deduction, and uncertainty. With the principled integration in Bayesian deep learning, the perception task and inference task are regarded as a whole and can benefit from each other. Concretely, being able to see the medical image could help with the doctor's diagnosis and inference. On the other hand, diagnosis and inference can, in turn, help understand the medical image. Suppose the doctor may not be sure about what a dark spot in a medical image is, but if she is able to \emph{infer} the etiology of the symptoms and disease, it can help her better decide whether the dark spot is a tumor or not. 

Take recommender systems \cite{CDL,DBLP:conf/aaai/LuDLX015,DBLP:journals/tkde/AdomaviciusK12,ricci2011introduction,DBLP:conf/recsys/LiuMLY11} as another example. A highly accurate recommender system requires (1) thorough understanding of item content (e.g., content in documents and movies) \cite{DBLP:journals/tkde/Park13}, (2) careful analysis of users' profiles/preferences \cite{DBLP:journals/tkde/WeiMJ05,DBLP:journals/tkde/YapTP07,DBLP:conf/aaai/ZhengCZXY10}, and (3) proper evaluation of similarity among users \cite{DBLP:journals/tkde/CaiLLMTL14,DBLP:journals/tkde/TangQZX13,DBLP:journals/tkde/HornickT12,DBLP:journals/tkde/BartoliniZP11}. 
% As another example, to achieve high accuracy in recommender systems \cite{CDL,DBLP:conf/aaai/LuDLX015,DBLP:journals/tkde/AdomaviciusK12,ricci2011introduction,DBLP:conf/recsys/LiuMLY11}, we need to fully understand the content of items (e.g., documents and movies) \cite{DBLP:journals/tkde/Park13}, analyze the profile and preference of users \cite{DBLP:journals/tkde/WeiMJ05,DBLP:journals/tkde/YapTP07,DBLP:conf/aaai/ZhengCZXY10}, and evaluate the similarity among users \cite{DBLP:journals/tkde/CaiLLMTL14,DBLP:journals/tkde/TangQZX13,DBLP:journals/tkde/HornickT12,DBLP:journals/tkde/BartoliniZP11}.
Deep learning with its ability to efficiently process dense high-dimensional data such as movie content is good at the first subtask, while PGM specializing in modeling conditional dependencies among users, items, and ratings (see Figure~\ref{fig:cdl_pgm} as an example, where $\u$, $\v$, and $\R$ are user latent vectors, item latent vectors, and ratings, respectively) excels at the other two. Hence unifying them two in a single principled probabilistic framework gets us the best of both worlds. Such integration also comes with additional benefit that uncertainty in the recommendation process is handled elegantly. What's more, one can also derive Bayesian treatments for concrete models, leading to more robust predictions~\cite{CDL,ColVAE}. 
% Besides the fact that better understanding of item content would help with the analysis of user profiles, the estimated similarity among users could provide valuable information for understanding item content in return. In order to fully utilize this bidirectional effect to boost recommendation accuracy, one might wish to unify deep learning and PGM in a single principled probabilistic framework, as done in \cite{CDL}.

As a third example, consider controlling a complex dynamical system according to the live video stream received from a camera. 
This problem can be transformed into iteratively performing two tasks, perception from raw images and control based on dynamic models. The perception task of processing raw images can be handled by deep learning while the control task usually needs more sophisticated models such as hidden Markov models and Kalman filters \cite{harrison1999bayesian,DBLP:conf/uai/MatsubaraGK14}. The feedback loop is then completed by the fact that actions chosen by the control model can affect the received video stream in turn. To enable an effective iterative process between the perception task and the control task, we need information to flow back and forth between them. The perception component would be the basis on which the control component estimates its states and the control component with a dynamic model built in would be able to predict the future trajectory (images). Therefore Bayesian deep learning is a suitable choice \cite{watter2015embed} for this problem. Note that similar to the recommender system example, both noise from raw images and uncertainty in the control process can be naturally dealt with under such a probabilistic framework. 

The above examples demonstrate BDL's major advantages as a principled way of unifying deep learning and PGM: information exchange between the \emph{perception task} and the \emph{inference task}, conditional dependencies on high-dimensional data, and effective modeling of uncertainty. In terms of uncertainty, it is worth noting that when BDL is applied to complex tasks, there are \emph{three kinds of parameter uncertainty} that need to be taken into account:
\begin{compactenum}
 \item Uncertainty on the neural network parameters.
 \item Uncertainty on the task-specific parameters.
 \item Uncertainty of exchanging information between the perception component and the task-specific component.
\end{compactenum}
By representing the unknown parameters using distributions instead of point estimates, BDL offers a promising framework to handle these three kinds of uncertainty in a unified way. It is worth noting that the third uncertainty could only be handled under a unified framework like BDL; training the perception component and the task-specific component separately is equivalent to assuming no uncertainty when \emph{exchanging information} between them two. Note that neural networks are usually over-parameterized and therefore pose additional challenges in efficiently handling the uncertainty in such a large parameter space. On the other hand, graphical models are often more concise and have smaller parameter space, providing better interpretability. 

Besides the advantages above, another benefit comes from the implicit regularization built in BDL. By imposing a prior on hidden units, parameters defining a neural network, or the model parameters specifying the conditional dependencies, BDL can to some degree avoid overfitting, especially when we have insufficient data. Usually, a BDL model consists of two components, a \emph{perception component} that is a Bayesian formulation of a certain type of neural networks and a \emph{task-specific component} that describes the relationship among different hidden or observed variables using PGM. Regularization is crucial for them both. Neural networks are usually heavily over-parameterized and therefore needs to be regularized properly. Regularization techniques such as weight decay and dropout \cite{srivastava2014dropout} are shown to be effective in improving performance of neural networks and they both have Bayesian interpretations \cite{gal2015dropout}. In terms of the task-specific component, expert knowledge or prior information, as a kind of regularization, can be incorporated into the model through the prior we imposed to guide the model when data are scarce.

% Yet another advantage of using BDL for complex tasks (tasks that need both perception and inference) is that it provides a principled Bayesian approach of handling parameter uncertainty. When BDL is applied to complex tasks, there are \emph{three kinds of parameter uncertainty} that need to be taken into account:
% \begin{compactenum}
%  \item Uncertainty on the neural network parameters.
%  \item Uncertainty on the task-specific parameters.
%  \item Uncertainty of exchanging information between the perception component and the task-specific component.
% \end{compactenum}
% By representing the unknown parameters using distributions instead of point estimates, BDL offers a promising framework to handle these three kinds of uncertainty in a unified way. It is worth noting that the third uncertainty could only be handled under a unified framework like BDL. If we train the perception component and the task-specific component separately, it is equivalent to assuming no uncertainty when \emph{exchanging information} between the two components.

There are also challenges when applying BDL to real-world tasks. (1) First, it is nontrivial to design an efficient Bayesian formulation of neural networks with reasonable time complexity. This line of work is pioneered by \cite{mackay1992practical,hinton1993keeping,neal1995bayesian}, but it has not been widely adopted due to its lack of scalability. Fortunately, some recent advances in this direction \cite{DBLP:conf/nips/Graves11,kingma2013auto,DBLP:conf/icml/Hernandez-Lobato15b,DBLP:conf/icml/BlundellCKW15,balan2015bayesian,CDL,NPN} seem to shed light\footnote{In summary, reduction in time complexity can be achieved via expectation propagation~\cite{DBLP:conf/icml/Hernandez-Lobato15b}, the reparameterization trick~\cite{kingma2013auto,DBLP:conf/icml/BlundellCKW15}, probabilistic formulation of neural networks with maximum a posteriori estimates~\cite{CDL}, approximate variational inference with natural-parameter networks~\cite{NPN}, knowledge distillation~\cite{balan2015bayesian}, etc. We refer readers to \cite{NPN} for a detailed overview.} on the practical adoption of Bayesian neural network\footnote{Here we refer to the Bayesian treatment of neural networks as Bayesian neural networks. The other term, Bayesian deep learning, is retained to refer to complex Bayesian models with both a perception component and a task-specific component. See Section~\ref{sec:history} for a detailed discussion.}. (2) The second challenge is to ensure efficient and effective information exchange between the perception component and the task-specific component. Ideally both the first-order and second-order information (e.g., the mean and the variance) should be able to flow back and forth between the two components. A natural way is to represent the perception component as a PGM and seamlessly connect it to the task-specific PGM, as done in \cite{RSDAE,CDL,DPFA}.

This survey provides a comprehensive overview of BDL with concrete models for various applications. The rest of the survey is organized as follows: In Section \ref{sec:dl}, we provide a review of some basic deep learning models. Section~\ref{sec:pgm} covers the main concepts and techniques for PGM. These two sections serve as the preliminaries for BDL, and the next section, Section \ref{sec:bdl}, demonstrates the rationale for the unified BDL framework and details various choices for implementing its \emph{perception component} and \emph{task-specific component}. Section~\ref{sec:app} reviews the BDL models applied to various areas such as recommender systems, topic models, and control, showcasing how BDL works in supervised learning, unsupervised learning, and general representation learning, respectively. Section \ref{sec:summary} discusses some future research issues and concludes the paper.

\section{Deep Learning}\label{sec:dl}
Deep learning normally refers to neural networks with more than two layers. To better understand deep learning, here we start with the simplest type of neural networks, multilayer perceptrons (MLP), as an example to show how conventional deep learning works. After that, we will review several other types of deep learning models based on MLP.

\subsection{Multilayer Perceptrons}
Essentially a multilayer perceptron is a sequence of parametric nonlinear transformations. Suppose we want to train a multilayer perceptron to perform a regression task which maps a vector of $M$ dimensions to a vector of $D$ dimensions. We denote the input as a matrix $\X_0$ ($0$ means it is the $0$-th layer of the perceptron). The $j$-th row of $\X_0$, denoted as $\X_{0,j*}$, is an $M$-dimensional vector representing one data point. The target (the output we want to fit) is denoted as $\Y$. Similarly $\Y_{j*}$ denotes a $D$-dimensional row vector. The problem of learning an $L$-layer multilayer perceptron can be formulated as the following optimization problem:

\begin{align*}
\min\limits_{\{\W_l\},\{\b_l\}} ~&\|\X_L-\Y\|_F+\lambda\sum\limits_l\|\W_l\|_F^2\\
\mbox{subject to}~~&\X_{l}=\sigma(\X_{l-1}\W_l+\b_l), l=1,\dots,L-1\\
&\X_{L}=\X_{L-1}\W_L+\b_L,
\end{align*}
where $\sigma(\cdot)$ is an element-wise sigmoid function for a matrix and $\sigma(x)=\frac{1}{1+\exp(-x)}$. $\|\cdot\|_F$ denotes the Frobenius norm. The purpose of imposing $\sigma(\cdot)$ is to allow nonlinear transformation. Normally other transformations like $\tanh(x)$ and $\max(0,x)$ can be used as alternatives of the sigmoid function.

Here $\X_l$ ($l=1,2,\dots,L-1$) is the hidden units. As we can see, $\X_L$ can be easily computed once $\X_0$, $\W_l$, and $\b_l$ are given. Since $\X_0$ is given as input, one only needs to learn $\W_l$ and $\b_l$ here. Usually this is done using backpropagation and stochastic gradient descent (SGD). The key is to compute the gradients of the objective function with respect to $\W_l$ and $\b_l$. Denoting the value of the objective function as $E$, one can compute the gradients using the chain rule as:
\begin{align}
\frac{\partial E}{\partial \X_L}&=2(\X_L-\Y), \;\;\;\;
\frac{\partial E}{\partial \X_l}=(\frac{\partial E}{\partial \X_{l+1}}\circ\X_{l+1}\circ(1-\X_{l+1}))\W_{l+1}, \nonumber\\
\frac{\partial E}{\partial \W_l}&=\X_{l-1}^T(\frac{\partial E}{\partial \X_l}\circ\X_l\circ(1-\X_l)), \;\;\;\;
\frac{\partial E}{\partial \b_l}=mean(\frac{\partial E}{\partial \X_l}\circ\X_l\circ(1-\X_l),1),\nonumber
\end{align}
where $l=1,\dots,L$ and the regularization terms are omitted. $\circ$ denotes the element-wise product and $mean(\cdot,1)$ is the matlab operation on matrices. In practice, we only use a small part of the data (e.g., $128$ data points) to compute the gradients for each update. This is called stochastic gradient descent.

As we can see, in conventional deep learning models, only $\W_l$ and $\b_l$ are free parameters, which we will update in each iteration of the optimization. $\X_l$ is not a free parameter since it can be computed exactly if $\W_l$ and $\b_l$ are given.

\begin{figure}[!tb]
\begin{center}
%\framebox[4.0in]{$\;$}
%\includegraphics[height=5cm]{likeli1.eps}
\subfigure{
\includegraphics[height=3.5cm]{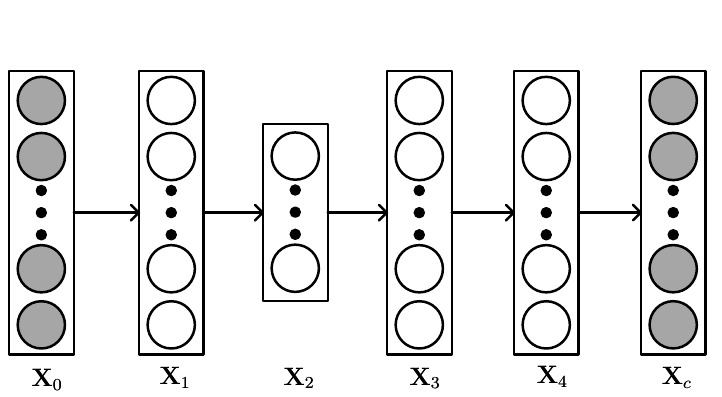}}
\hspace{0.5in}
\subfigure{
\includegraphics[height=3.0cm]{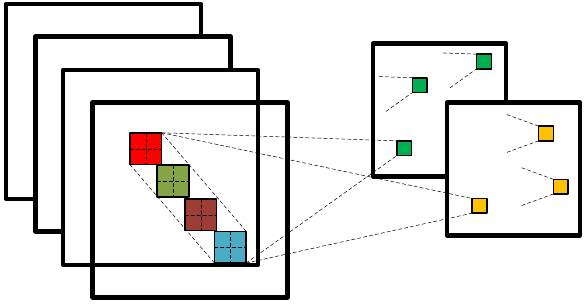}}
\end{center}
\vskip -0.2in
\caption{Left: A 2-layer SDAE with $L=4$. Right: A convolutional layer with $4$ input feature maps and $2$ output feature maps.
%$P=1$ is the sparse setting and $P=10$ is the dense setting.  $P$ is the number of ratings for each user in the training set.
}
\label{fig:sdae_ctr}
\vskip -0.2in
\end{figure}

\subsection{Autoencoders}\label{sec:ae}
An autoencoder (AE) is a feedforward neural network to encode the input into a more compact representation and reconstruct the input with the learned representation. In its simplest form, an autoencoder is no more than a multilayer perceptron with a bottleneck layer (a layer with a small number of hidden units) in the middle. The idea of autoencoders has been around for decades \cite{lecun-87,bourlard1988auto,hinton1994autoencoders,dlbook} and abundant variants of autoencoders have been proposed to enhance representation learning including sparse AE \cite{poultney2006efficient}, contrastive AE \cite{rifai2011contractive}, and denoising AE \cite{DBLP:journals/jmlr/VincentLLBM10}. For more details, please refer to a nice recent book on deep learning \cite{dlbook}. Here we introduce a kind of multilayer denoising AE, known as stacked denoising autoencoders (SDAE), both as an example of AE variants and as background for its applications on BDL-based recommender systems in Section \ref{sec:bdl}.

SDAE \cite{DBLP:journals/jmlr/VincentLLBM10} is a feedforward neural network for learning representations (encoding) of the input data by learning to predict the clean input itself in the output, as shown in Figure \ref{fig:sdae_ctr}(left). The hidden layer in the middle, i.e., $\X_2$ in the figure, can be constrained to be a bottleneck to learn compact representations. The difference between traditional AE and SDAE is that the input layer $\X_0$ is a \emph{corrupted} version of the \emph{clean} input data $\X_c$. Essentially an SDAE solves the following optimization problem:
%\footnote{*** Use $\|$ instead of $||$ for the norm. (** Modified.)}
\begin{align*}
\min\limits_{\{\W_l\},\{\b_l\}}& \|\X_c-\X_L\|_F^2+\lambda\sum\limits_l \|\W_l\|_F^2\\
\mbox{subject to}~~&\X_{l}=\sigma(\X_{l-1}\W_l+\b_l), l=1,\dots,L-1\\
&\X_{L}=\X_{L-1}\W_L+\b_L,
\end{align*}
where $\lambda$ is a regularization parameter. Here SDAE can be regarded as a multilayer perceptron for regression tasks described in the previous section. The input $\X_0$ of the MLP is the corrupted version of the data and the target $\Y$ is the clean version of the data $\X_c$. For example, $\X_c$ can be the raw data matrix, and we can randomly set $30\%$ of the entries in $\X_c$ to $0$ and get $\X_0$. In a nutshell, SDAE learns a neural network that takes the noisy data as input and recovers the clean data in the last layer. This is what `denoising' in the name means. Normally, the output of the middle layer, i.e., $\X_2$ in Figure \ref{fig:sdae_ctr}(left), would be used to compactly represent the data.

% \begin{figure}[!tb]
% \begin{center}
% %\framebox[4.0in]{$\;$}
% %\includegraphics[height=5cm]{likeli1.eps}
% \subfigure{
% \includegraphics[height=3.0cm]{fig/convolutional_layer}}
% \end{center}
% \vskip -0.2in
% \caption{A convolutional layer with $4$ input feature maps and $2$ output feature maps.
% %$P=1$ is the sparse setting and $P=10$ is the dense setting.  $P$ is the number of ratings for each user in the training set.
% }
% \label{fig:conv}
% \vskip -0.2in
% \end{figure}

\subsection{Convolutional Neural Networks}\label{sec:cnn}
Convolutional neural networks (CNN) can be viewed as another variant of MLP. Different from AE, which is initially designed to perform dimensionality reduction, CNN is biologically inspired. According to \cite{hubel1968receptive}, two types of cells have been identified in the cat's visual cortex. One is simple cells that respond maximally to specific patterns within their receptive field, and the other is complex cells with larger receptive field that are considered locally invariant to positions of patterns. Inspired by these findings, the two key concepts in CNN are then developed: convolution and max-pooling.

\textbf{Convolution}: In CNN, a feature map is the result of the convolution of the input and a linear filter, followed by some element-wise nonlinear transformation. The \emph{input} here can be the raw image or the feature map from the previous layer. Specifically, with input $\X$, weights $\W^k$, bias $b^k$, the $k$-th feature map $\H^k$ can be obtained as follows:
\begin{align*}
\H_{ij}^k=\tanh((\W^k*\X)_{ij}+b^k).
\end{align*}
Note that in the equation above we assume one single input feature map and multiple output feature maps. In practice, CNN often has multiple input feature maps as well due to its deep structure. A convolutional layer with $4$ input feature maps and $2$ output feature maps is shown in Figure \ref{fig:sdae_ctr}(right).

\textbf{Max-Pooling}: Traditionally, a convolutional layer in CNN is followed by a max-pooling layer, which can be seen as a type of nonlinear downsampling. The operation of max-pooling is simple. For example, if we have a feature map of size $6\times 9$, the result of max-pooling with a $3\times 3$ region would be a downsampled feature map of size $2\times 3$. Each entry of the downsampled feature map is the maximum value of the corresponding $3\times 3$ region in the $6\times 9$ feature map. Max-pooling layers can not only reduce computational cost by ignoring the non-maximal entries but also provide local translation invariance.

\textbf{Putting it all together}: Usually to form a complete and working CNN, the input would alternate between convolutional layers and max-pooling layers before going into an MLP for tasks such as classification or regression. One classic example is the LeNet-5 \cite{lecun1998gradient}, which alternates between $2$ convolutional layers and $2$ max-pooling layers before going into a fully connected MLP for target tasks.

\begin{figure}[!tb]
\begin{center}
%\framebox[4.0in]{$\;$}
%\includegraphics[height=5cm]{likeli1.eps}
\subfigure{
\includegraphics[height=2.6cm]{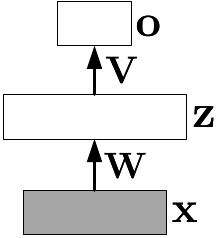}}
\hspace{0.5in}
\subfigure{
\includegraphics[height=2.6cm]{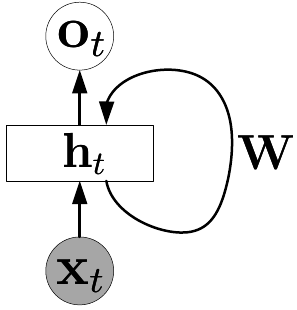}}
\hspace{0.5in}
\subfigure{
\includegraphics[height=2.6cm]{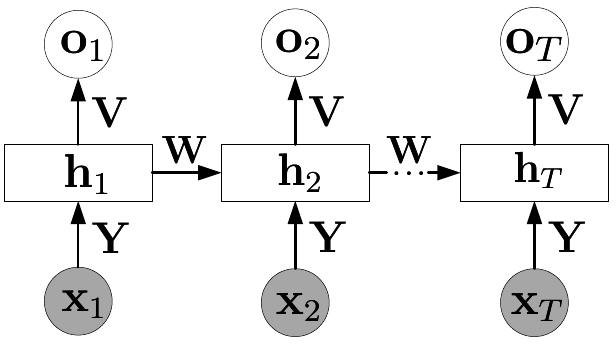}}
\end{center}
\vskip -0.2in
\caption{Left: A conventional feedforward neural network with one hidden layer, where $\x$ is the input, $\z$ is the hidden layer, and $\oo$ is the output, $\W$ and $\V$ are the corresponding weights (biases are omitted here). Middle: A recurrent neural network with input $\{\x_t\}_{t=1}^T$, hidden states $\{\h_t\}_{t=1}^T$, and output $\{\oo_t\}_{t=1}^T$. Right: An unrolled RNN which is equivalent to the one in Figure \ref{fig:ffn_rnn}(middle). Here each node (e.g., $\x_1$, $\h_1$, or $\oo_1$) is associated with one particular time step.
}
\label{fig:ffn_rnn}
\vskip -0.2in
\end{figure}

% \begin{figure}[!tb]
% \begin{center}
% \includegraphics[height=2.4cm]{fig/rrnet_unroll}
% \end{center}
% \vskip -0.2in
% \caption{An unrolled RNN which is equivalent to the one in Figure \ref{fig:ffn_rnn}(right). Here each node (e.g., $\x_1$, $\h_1$, or $\oo_1$) is associated with one particular time instance.
% }
% \label{fig:rnn_unroll}
% \vskip -0in
% \end{figure}

\subsection{Recurrent Neural Network}\label{sec:rnn}
When reading an article, one normally takes in one word at a time and try to understand the current word based on previous words. This is a recurrent process that needs short-term memory. Unfortunately conventional feedforward neural networks like the one shown in Figure \ref{fig:ffn_rnn}(left) fail to do so. For example, imagine we want to constantly predict the next word as we read an article. Since the feedforward network only computes the output $\oo$ as $\V q(\W \x)$, where the function $q(\cdot)$ denotes element-wise nonlinear transformation, it is unclear how the network could naturally model the sequence of words to predict the next word.

\subsubsection{Vanilla Recurrent Neural Network}

To solve the problem, we need a recurrent neural network~\cite{dlbook} instead of a feedforward one. As shown in Figure \ref{fig:ffn_rnn}(middle),
%\footnote{*** Redraw the diagram to align the two arrows for the input-to-hidden and hidden-to-output weights.}
the computation of the current hidden states $\h_t$ depends on the current input $\x_t$ (e.g., the $t$-th word) and the previous hidden states $\h_{t-1}$. This is why there is a loop in the RNN. It is this loop that enables short-term memory in RNNs. The $\h_t$ in the RNN represents what the network knows so far at the $t$-th time step. To see the computation more clearly, we can unroll the loop and represent the RNN as in Figure \ref{fig:ffn_rnn}(right). If we use hyperbolic tangent nonlinearity ($\tanh$), the computation of output $\oo_t$ will be as follows:
\begin{align*}
\a_t = \W\h_{t-1}+\Y\x_t+\b, \;\;\;\;\;
\h_t = \tanh(\a_t), \;\;\;\;\;
\oo_t = \V\h_t+\c,
\end{align*}
where $\Y$, $\W$, and $\V$ denote the weight matrices for input-to-hidden, hidden-to-hidden, and hidden-to-output connections, respectively, and $\b$ and $\c$ are the corresponding biases. If the task is to classify the input data at each time step, we can compute the classification probability as $\p_t=\mbox{softmax}(\oo_t)$ where
\begin{align*}
\mbox{softmax}(\q) = \frac{\exp(\q)}{\sum\limits_i \exp(\q_i)}.
\end{align*}

\begin{figure*}[!tb]
\begin{center}
%\framebox[4.0in]{$\;$}
%\includegraphics[height=5cm]{likeli1.eps}
\includegraphics[height=1.5cm]{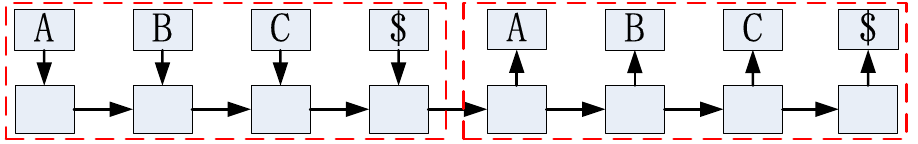}
\end{center}
\vskip -0.2in
\caption{The encoder-decoder architecture involving two LSTMs. The encoder LSTM (in the left rectangle) encodes the sequence `ABC' into a representation and the decoder LSTM (in the right rectangle) recovers the sequence from the representation. `\$' marks the end of a sentence.
}
\label{fig:lstmae}
\vskip -0.2in
\end{figure*}

Similar to feedforward networks, an RNN is trained with a generalized back-propagation algorithm called \emph{back-propagation through time} (BPTT) \cite{dlbook}. Essentially the gradients are computed through the unrolled network as shown in Figure \ref{fig:ffn_rnn}(right) with shared weights and biases for all time steps. 

\subsubsection{Gated Recurrent Neural Network}\label{sec:lstm}
The problem with the vanilla RNN above is that the gradients propagated over many time steps are prone to vanish or explode, making the optimization notoriously difficult. In addition, the signal passing through the RNN decays exponentially, making it impossible to model long-term dependencies in long sequences. Imagine we want to predict the last word in the paragraph `I have many books ... I like \emph{reading}'. In order to get the answer, we need `long-term memory' to retrieve information (the word `books') at the start of the text. To address this problem, the long short-term memory model (LSTM) is designed as a type of gated RNN to model and accumulate information over a relatively long duration. The intuition behind LSTM is that when processing a sequence consisting of several subsequences, it is sometimes useful for the neural network to summarize or forget the old states before moving on to process the next subsequence \cite{dlbook}.
%In our setting, the LSTM takes a sequence of $T_j$ words (corresponding to the $j$-th item) as input. Each word is encoded in a $1$-of-$S$ representation, that is, a vector of length $S$ with all zeros except for the element corresponding to the word.
%\footnote{*** Already described above. ** Deleted.}
Using $t=1\dots T_j$ to index the words in the sequence, the formulation of LSTM is as follows (we drop the item index $j$ for notational simplicity):
\begin{align}
\x_t = \W_w\e_t, \;\;\;\;\;
\s_{t} = \h_{t-1}^f\odot\s_{t-1}+\h_{t-1}^i\odot\sigma(\Y\x_{t-1}+\W\h_{t-1}+\b) \label{eq:gate},
\end{align}
where $\x_t$ is the word embedding of the $t$-th word, $\W_w$ is a $K_W$-by-$S$ word embedding matrix, and $\e_t$ is the $1$-of-$S$ representation, $\odot$ stands for the element-wise product operation between two vectors, $\sigma(\cdot)$ denotes the sigmoid function, $\s_t$ is the cell state of the $t$-th word, and $\b$, $\Y$, and $\W$ denote the biases, input weights, and recurrent weights respectively. The forget gate units $\h_t^f$ and the input gate units $\h_t^i$ in Equation (\ref{eq:gate}) can be computed using their corresponding weights and biases $\Y^f$, $\W^f$, $\Y^i$, $\W^i$, $\b^f$, and $\b^i$:
\begin{align*}
\h_t^f = \sigma(\Y^f\x_t+\W^f\h_t+\b^f), \;\;\;\;\;
\h_t^i = \sigma(\Y^i\x_t+\W^i\h_t+\b^i).
\end{align*}
The output depends on the output gate $\h_t^o$ which has its own weights and biases $\Y^o$, $\W^o$, and $\b^o$:
\begin{align*}
\h_t = \tanh(\s_{t})\odot\h_{t-1}^o, \;\;\;\;\;
\h_t^o = \sigma(\Y^o\x_t+\W^o\h_t+\b^o).
\end{align*}
Note that in the LSTM, information of the processed sequence is contained in the cell states $\s_t$ and the output states $\h_t$, both of which are column vectors of length $K_W$.

Similar to \cite{sutskever2014sequence,DBLP:conf/emnlp/ChoMGBBSB14}, we can use the output state and cell state at the last time step ($\h_{T_j}$ and $\s_{T_j}$) of the first LSTM as the initial output state and cell state of the second LSTM. This way the two LSTMs can be concatenated to form an encoder-decoder architecture, as shown in Figure \ref{fig:lstmae}.

Note that there is a vast literature on deep learning and neural networks. The introduction in this section intends to serve only as the background of Bayesian deep learning. Readers are referred to \cite{dlbook} for a comprehensive survey and more details.
%Note that besides the deep learning models discussed above, there are many others, including restricted Boltzmann machine \cite{RBM}, sigmoid belief networks \cite{neal1992}, deep belief networks \cite{DBN}, and so on. Readers are referred to \cite{dlbook} for a comprehensive survey and more details.

\section{Probabilistic Graphical Models}\label{sec:pgm} %tocite
Probabilistic Graphical Models (PGM) use diagrammatic representations to describe random variables and relationships among them. Similar to a graph that contains nodes (vertices) and links (edges), PGM has nodes to represent random variables and links to indicate probabilistic relationships among them.

\subsection{Models}
There are essentially two types of PGM, directed PGM (also known as Bayesian networks) and undirected PGM (also known as Markov random fields)~\cite{PRML}. In this survey we mainly focus on directed PGM\footnote{For convenience, PGM stands for directed PGM in this survey unless specified otherwise.}. For details on undirected PGM, readers are referred to \cite{PRML}.

\begin{figure*}[!tb]
\begin{center}
\includegraphics[height=1.5cm]{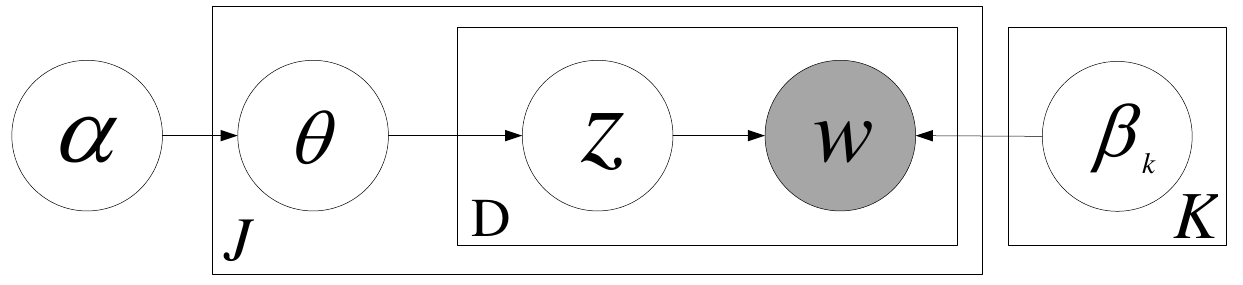}
\end{center}
\vskip -0.2in
\caption{The probabilistic graphical model for LDA, $J$ is the number of documents, $D$ is the number of words in a document, and $K$ is the number of topics.
}
\label{fig:lda}
\vskip -0.1in
\end{figure*}

A classic example of PGM would be latent Dirichlet allocation (LDA), which is used as a topic model to analyze the generation of words and topics in documents~\cite{LDA}. Usually PGM comes with a graphical representation of the model and a generative process to depict the story of how the random variables are generated step by step. Figure \ref{fig:lda} shows the graphical model for LDA and the corresponding generative process is as follows:
\begin{itemize}
\item For each document $j$ ($j=1,2,\dots,J$),
\begin{enumerate}
\item Draw topic proportions $\theta_j\sim\mbox{Dirichlet}(\alpha)$.
\item For each word $w_{jn}$ of item~(document) $\w_j$,
\begin{enumerate}
\item Draw topic assignment $z_{jn}\sim\mbox{Mult}(\theta_j)$.
\item Draw word $w_{jn}\sim\mbox{Mult}(\beta_{z_{jn}})$.
\end{enumerate}
\end{enumerate}
\end{itemize}

The generative process above provides the story of how the random variables are generated. In the graphical model in Figure \ref{fig:lda}, the shaded node denotes observed variables while the others are latent variables ($\tha$ and $\z$) or parameters ($\alpha$ and $\beta$). Once the model is defined, learning algorithms can be applied to automatically learn the latent variables and parameters.

Due to its Bayesian nature, PGM such as LDA is easy to extend to incorporate other information or to perform other tasks. For example, following LDA, different variants of topic models have been proposed. \cite{DTM,cDTM} are proposed to incorporate temporal information, and \cite{CTM} extends LDA by assuming correlations among topics. \cite{onlineLDA} extends LDA from the batch mode to the online setting, making it possible to process large datasets. On recommender systems, collaborative topic regression (CTR)~\cite{CTR} extends LDA to incorporate rating information and make recommendations. This model is then further extended to incorporate social information \cite{CTR-SMF,CTRSR,RCTR}.

\begin{table*}[!t]
\newcommand{\tabincell}[2]{\begin{tabular}{@{}#1@{}}#2\end{tabular}}
\caption{\small Summary of BDL Models with Different Learning Algorithms (MAP: Maximum a Posteriori, VI: Variational Inference, Hybrid MC: Hybrid Monte Carlo) and Different Variance Types (ZV: Zero-Variance, HV: Hyper-Variance, LV: Learnable-Variance). 
}\label{table:summary}
\begin{center}
\vskip -0.3cm
\begin{scriptsize}
% \begin{scriptsize}
\begin{tabular}{c|l|c|cccc}
\hline
 Applications & Models & Variance of $\Om_h$ & MAP & VI & Gibbs Sampling & Hybrid MC \tabularnewline
\hline

\multirow{8}{*}{\tabincell{c}{Recommender\\Systems}} & Collaborative Deep Learning (CDL)~\cite{CDL} &  HV & \checkmark  & & & \tabularnewline
% \cline{2-7}
 & Bayesian CDL~\cite{CDL} &  HV & & & \checkmark &  \tabularnewline
% \cline{2-7}
& Marginalized CDL~\cite{li2015deep} &  LV & \checkmark & & &  \tabularnewline
% \cline{2-7}
& Symmetric CDL~\cite{li2015deep} &  LV & \checkmark & &  &  \tabularnewline
% \cline{2-7}
& Collaborative Deep Ranking~\cite{yingcollaborative} &  HV & \checkmark & &  &  \tabularnewline
% \cline{2-7}
& Collaborative Knowledge Base Embedding~\cite{CKE} &  HV & \checkmark & &  &  \tabularnewline
% \cline{2-7}
& Collaborative Recurrent AE~\cite{CRAE}  & HV & \checkmark & &  &  \tabularnewline
% \cline{2-7}
& Collaborative Variational Autoencoders~\cite{ColVAE} &  HV &  & \checkmark &  &  \tabularnewline
\hline

\multirow{4}{*}{\tabincell{l}{\\~Topic\\Models}} & Relational SDAE &  HV & \checkmark  & & & \tabularnewline
% \cline{2-7}
 & Deep Poisson Factor Analysis with Sigmoid Belief Networks~\cite{DPFA} &  ZV & & & \checkmark & \checkmark \tabularnewline
% \cline{2-7}
& Deep Poisson Factor Analysis with Restricted Boltzmann Machine~\cite{DPFA} &  ZV &  & & \checkmark & \checkmark \tabularnewline
% \cline{2-7}
& Deep Latent Dirichlet Allocation~\cite{DLDA} &  LV &  & &  & \checkmark \tabularnewline
% \cline{2-7}
& Dirichlet Belief Networks~\cite{DirBN} &  LV &  & & \checkmark &  \tabularnewline
\hline

\multirow{4}{*}{\tabincell{l}{Control}} & Embed to Control~\cite{watter2015embed} &  LV &   & \checkmark & & \tabularnewline
% \cline{2-7}
& Deep Variational Bayes Filters~\cite{DVBF} &  LV &   & \checkmark & & \tabularnewline
% \cline{2-7}
& Probabilistic Recurrent State-Space Models~\cite{PR-SSM} &  LV &   & \checkmark & & \tabularnewline
% \cline{2-7}
& Deep Planning Networks~\cite{PlaNet} &  LV &   & \checkmark & & \tabularnewline
\hline

\multirow{3}{*}{\tabincell{c}{Link\\Prediction}} & Relational Deep Learning~\cite{RDL} &  LV & \checkmark  & \checkmark & & \tabularnewline
% \cline{2-7}
 & Graphite~\cite{Graphite} &  LV & & \checkmark &  &  \tabularnewline
% \cline{2-7}
& Deep Generative Latent Feature Relational Model~\cite{SBGNN} &  LV &  & \checkmark &  &  \tabularnewline
\hline

\multirow{2}{*}{\tabincell{l}{NLP}} & Sequence to Better Sequence~\cite{S2BS} &  LV &   & \checkmark & & \tabularnewline
% \cline{2-7}
& Quantifiable Sequence Editing~\cite{QuaSE} &  LV &   & \checkmark & & \tabularnewline
\hline

\multirow{4}{*}{\tabincell{c}{Computer\\Vision}} & Asynchronous Temporal Fields~\cite{ATF} &  LV &   & \checkmark & & \tabularnewline
% \cline{2-7}
 & Attend, Infer, Repeat (AIR)~\cite{AIR} &  LV & & \checkmark &  &  \tabularnewline
% \cline{2-7}
& Fast AIR~\cite{FastAIR} &  LV &  & \checkmark &  &  \tabularnewline
% \cline{2-7}
& Sequential AIR~\cite{SQAIR} &  LV &  & \checkmark &  &  \tabularnewline
\hline

\multirow{5}{*}{\tabincell{c}{Speech}} & Factorized Hierarchical VAE~\cite{FHVAE} &  LV &   & \checkmark & & \tabularnewline
% \cline{2-7}
 & Scalable Factorized Hierarchical VAE~\cite{SFHVAE} &  LV & & \checkmark &  &  \tabularnewline
% \cline{2-7}
& Gaussian Mixture Variational Autoencoders~\cite{GMVAE} &  LV &  & \checkmark &  &  \tabularnewline
% \cline{2-7}
& Recurrent Poisson Process Units~\cite{RPPU} &  LV & \checkmark & \checkmark &  &  \tabularnewline
& Deep Graph Random Process~\cite{DGP} &  LV & \checkmark & \checkmark &  &  \tabularnewline
\hline

\multirow{4}{*}{\tabincell{c}{Time Series\\Forecasting}} & DeepAR~\cite{DeepAR} &  LV &   & \checkmark & & \tabularnewline
% \cline{2-7}
 & DeepState~\cite{DeepState} &  LV & & \checkmark &  &  \tabularnewline
% \cline{2-7}
& Spline Quantile Function RNN~\cite{SQF-RNN} &  LV &  & \checkmark &  &  \tabularnewline
% \cline{2-7}
& DeepFactor~\cite{DeepFactor} &  LV & & \checkmark &  &  \tabularnewline
\hline

\multirow{4}{*}{\tabincell{c}{Health\\Care}} & Deep Poisson Factor Models~\cite{DPFM} &  LV &  &  & & \checkmark \tabularnewline
% \cline{2-7}
 & Deep Markov Models~\cite{DeepMarkov} &  LV & & \checkmark &  &  \tabularnewline
% \cline{2-7}
& Black-Box False Discovery Rate~\cite{BBFDR} &  LV &  & \checkmark &  &  \tabularnewline
% \cline{2-7}
& Bidirectional Inference Networks~\cite{BIN} &  LV & \checkmark & &  &  \tabularnewline
\hline

\hline
\end{tabular}
%\end{sc}
\end{scriptsize}
% \end{scriptsize}
\end{center}
\vskip -0.2in
\end{table*}

\subsection{Inference and Learning}
Strictly speaking, the process of finding the parameters (e.g., $\alpha$ and $\beta$ in Figure \ref{fig:lda}) is called learning and the process of finding the latent variables (e.g., $\tha$ and $\z$ in Figure \ref{fig:lda}) given the parameters is called inference. However, given only the observed variables (e.g. $\w$ in Figure \ref{fig:lda}), learning and inference are often intertwined. Usually the learning and inference of LDA would alternate between the updates of latent variables (which correspond to inference) and the updates of the parameters (which correspond to learning). Once the learning and inference of LDA is completed, one could obtain the learned parameters $\alpha$ and $\beta$. If a new document comes, one can now fix the learned $\alpha$ and $\beta$ and then perform inference alone to find the topic proportions $\theta_j$ of the new document.\footnote{For convenience, we use `learning' to represent both `learning and inference' in the following text.}

Similar to LDA, various learning and inference algorithms are available for each PGM. Among them, the most cost-effective one is probably maximum a posteriori (MAP), which amounts to maximizing the posterior probability of the latent variable. Using MAP, the learning process is equivalent to minimizing (or maximizing) an objective function with regularization. One famous example is the probabilistic matrix factorization (PMF) \cite{PMF}, where the learning of the graphical model is equivalent to factorizing a large matrix into two low-rank matrices with L2 regularization.

MAP, as efficient as it is, gives us only \emph{point estimates} of latent variables (and parameters). In order to take the uncertainty into account and harness the full power of Bayesian models, one would have to resort to Bayesian treatments such as variational inference and Markov chain Monte Carlo (MCMC). For example, the original LDA uses variational inference to approximate the true posterior with factorized variational distributions \cite{LDA}. Learning of the latent variables and parameters then boils down to minimizing the KL-divergence between the variational distributions and the true posterior distributions. Besides variational inference, another choice for a Bayesian treatment is MCMC. For example, MCMC algorithms such as \cite{porteous2008fast} have been proposed to learn the posterior distributions of LDA.

\section{Bayesian Deep Learning}\label{sec:bdl}
With the preliminaries on deep learning and PGM, we are now ready to introduce the general framework and some concrete examples of BDL. Specifically, in this section we will list some recent BDL models with applications on recommender systems, topic models, control, etc. A summary of these models is shown in Table \ref{table:summary}.

\subsection{A Brief History of Bayesian Neural Networks and Bayesian Deep Learning}\label{sec:history}
One topic highly related to BDL is Bayesian neural networks (BNN) or Bayesian treatments of neural networks. Similar to any Bayesian treatment, BNN imposes a prior on the neural network's parameters and aims to learn a posterior distribution of these parameters. During the inference phrase, such a distribution is then marginalized out to produce final predictions. In general such a process is called Bayesian model averaging~\cite{PRML} and can be seen as learning an infinite number of (or a distribution over) neural networks and then aggregating the results through ensembling. 

The study of BNN dates back to 1990s with notable works from~\cite{mackay1992practical,hinton1993keeping,neal1995bayesian}. Over the years, a large body of works~\cite{DBLP:conf/nips/Graves11,kingma2013auto,DBLP:conf/icml/Hernandez-Lobato15b,DBLP:conf/icml/BlundellCKW15,balan2015bayesian,MaxNPN} have emerged to enable substantially better scalability and incorporate recent advancements of deep neural networks. Due to BNN's long history, the term `Bayesian deep learning' sometimes specifically refers to `Bayesian neural networks'~\cite{maddox2019simple,BDL-report}. In this survey, we instead use `Bayesian deep learning' in a broader sense to refer to the probabilistic framework subsuming Bayesian neural networks. To see this, note that a BDL model with a \emph{perception component} and an empty \emph{task-specific component} is equivalent to a Bayesian neural network (details on these two components are discussed in Section~\ref{sec:general}). 

Interestingly, though BNN started in 1990s, the study of BDL in a broader sense started roughly in 2014~\cite{RSDAE,CDL,DPFM,BDL-thesis}, slightly after the deep learning breakthrough in the ImageNet LSVRC contest in 2012~\cite{krizhevsky2012imagenet}. As we will see in later sections, BNN is usually used as a perception component in BDL models. %BNN as a perception component TODO

Today BDL is gaining more and more popularity, has found successful applications in areas such as recommender systems and computer vision, and appears as the theme of various conference workshops (e.g., the NeurIPS BDL workshop\footnote{\url{http://bayesiandeeplearning.org/}}). %, and is listed as areas (ICML/NeurIPS)

% \textred{BNN started in 1990s while BDL in a broader sense started roughly in 2014, slightly after the 2012 ImageNet DL breakthrough in CV}

% \textred{difference between BNN and BDL}

\subsection{General Framework}\label{sec:general}
As mentioned in Section \ref{sec:intro}, BDL is a principled probabilistic framework with two seamlessly integrated components: a \emph{perception component} and a \emph{task-specific component}.

\begin{figure}[!tb]
\begin{center}
\includegraphics[height=2.5cm]{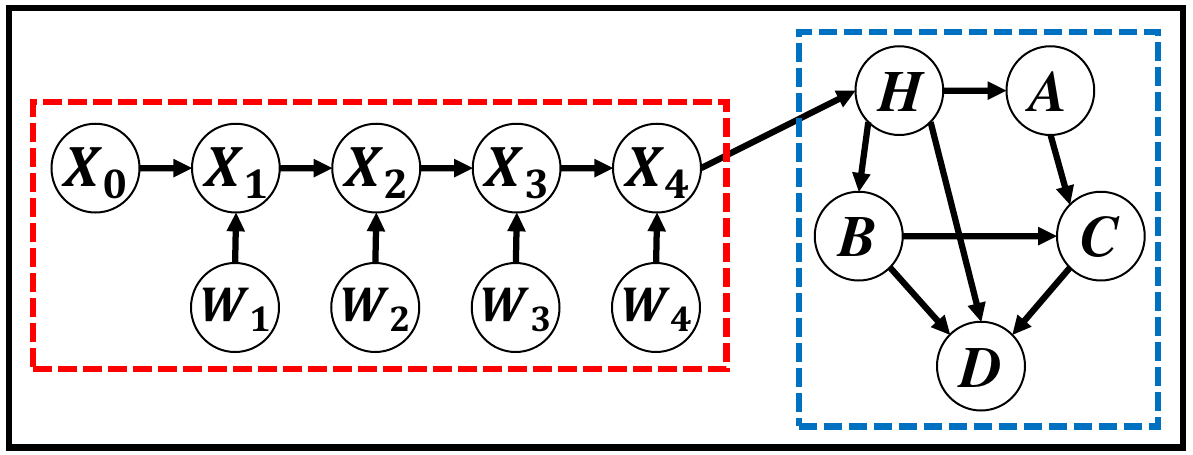}
\end{center}
\vskip -0.1in
\caption{The PGM for an example BDL. The red rectangle on the left indicates the perception component, and the blue rectangle on the right indicates the task-specific component. The hinge variable $\Om_h=\{\H\}$.
}
\label{fig:BDL_typeI}
\vskip -0.2in
\end{figure}

\textbf{Two Components}: Figure \ref{fig:BDL_typeI} shows the PGM of a simple BDL model as an example. The part inside the red rectangle on the left represents the perception component and the part inside the blue rectangle on the right is the task-specific component. Typically, the perception component would be a probabilistic formulation of a deep learning model with multiple nonlinear processing layers represented as a chain structure in the PGM. While the nodes and edges in the perception component are relatively simple, those in the task-specific component often describe more complex distributions and relationships among variables. Concretely, a task-specific component can take various forms. For example, it can be a typical Bayesian network (directed PGM) such as LDA, a deep Bayesian network~\cite{BIN}, or a stochastic process~\cite{ross1996stochastic,RPPU}, all of which can be represented in the form of PGM.

\textbf{Three Variable Sets}: There are three sets of variables in a BDL model: perception variables, hinge variables, and task variables. In this paper, we use $\Om_p$ to denote the set of perception variables (e.g., $\X_0$, $\X_1$, and $\W_1$ in Figure \ref{fig:BDL_typeI}), which are the variables in the perception component. Usually $\Om_p$ would include the weights and neurons in the probabilistic formulation of a deep learning model. $\Om_h$ is used to denote the set of hinge variables (e.g. $\H$ in Figure \ref{fig:BDL_typeI}). These variables directly interact with the perception component from the task-specific component. The set of task variables (e.g. $\A$, $\B$, and $\C$ in Figure \ref{fig:BDL_typeI}), i.e., variables in the task-specific component without direct relation to the perception component, is denoted as $\Om_t$.

\textbf{Generative Processes for Supervised and Unsupervised Learning}: If the edges between the two components \emph{point towards} $\Om_h$, the joint distribution of all variables can be written as:
\begin{align}
p(\Om_p,\Om_h,\Om_t)=p(\Om_p)p(\Om_h|\Om_p)p(\Om_t|\Om_h). \label{eq:typeone}
\end{align}

If the edges between the two components \emph{originate from} $\Om_h$, the joint distribution of all variables can be written as:
\begin{align}
p(\Om_p,\Om_h,\Om_t)=p(\Om_t)p(\Om_h|\Om_t)p(\Om_p|\Om_h). \label{eq:typetwo}
\end{align}

Equation~(\ref{eq:typeone}) and (\ref{eq:typetwo}) assume different generative processes for the data and correspond to different learning tasks. The former is usually used for supervised learning, where the perception component serves as a probabilistic (or Bayesian) representation learner to facilitate any downstream tasks (see Section~\ref{sec:recsys} for some examples). The latter is usually used for unsupervised learning, where the task-specific component provides structured constraints and domain knowledge to help the perception component learn stronger representations (see Section~\ref{sec:topic_models} for some examples). 

Note that besides these two vanilla cases, it is possible for BDL to simultaneously have some edges between the two components pointing towards $\Om_h$ and some originating from $\Om_h$, in which case the decomposition of the joint distribution would be more complex.

\textbf{Independence Requirement}:
The introduction of hinge variables $\Om_h$ and related conditional distributions simplifies the model (especially when $\Om_h$'s in-degree or out-degree is $1$), facilitate learning, and provides inductive bias to concentrate information inside $\Om_h$. Note that hinge variables are always in the task-specific component; the connections between hinge variables $\Om_h$ and the perception component (e.g., $\X_4\rightarrow \H$ in Figure \ref{fig:BDL_typeI}) should normally be independent for convenience of parallel computation in the perception component. For example, each row in $\H$ is related to only one corresponding row in $\X_4$. Although it is not mandatory in BDL models, meeting this requirement would significantly increase the efficiency of parallel computation in model training.

\textbf{Flexibility of Variance for $\Om_h$}: As mentioned in Section \ref{sec:intro}, one of BDL's motivations is to model the \emph{uncertainty of exchanging information} between the perception component and the task-specific component, which boils down to modeling the uncertainty related to $\Om_h$. For example, such uncertainty is reflected in the variance of the conditional density $p(\Om_h|\Om_p)$ in Equation (\ref{eq:typeone})\footnote{For models with the joint likelihood decomposed as in Equation (\ref{eq:typetwo}), the uncertainty is reflected in the variance of $p(\Om_p|\Om_h)$.}. According to the degree of flexibility, there are three types of variance for $\Om_h$ (for simplicity we assume the joint likelihood of BDL is Equation (\ref{eq:typeone}), $\Om_p=\{p\}$, $\Om_h=\{h\}$, and $p(\Om_h|\Om_p)=\NM(h|\mu_p,\sigma_p^2)$ in our example):
\begin{compactitem}
\item \textbf{Zero-Variance}: Zero-Variance (ZV) assumes no uncertainty during the information exchange between the two components. In the example, zero-variance means directly setting $\sigma_p^2$ to $0$.
\item \textbf{Hyper-Variance}: Hyper-Variance (HV) assumes that uncertainty during the information exchange is defined through hyperparameters. In the example, HV means that $\sigma_p^2$ is a manually tuned hyperparameter.
\item \textbf{Learnable Variance}: Learnable Variance (LV) uses learnable parameters to represent uncertainty during the information exchange. In the example, $\sigma_p^2$ is the learnable parameter.
\end{compactitem}
As shown above, we can see that in terms of model flexibility, $\text{LV}>\text{HV}>\text{ZV}$. Normally, if properly regularized, an LV model outperforms an HV model, which is superior to a ZV model. In Table \ref{table:summary}, we show the types of variance for $\Om_h$ in different BDL models. Note that although each model in the table has a specific type, one can always adjust the models to devise their counterparts of other types. For example, while CDL in the table is an HV model, we can easily adjust $p(\Om_h|\Om_p)$ in CDL to devise its ZV and LV counterparts. In \cite{CDL}, the authors compare the performance of an HV CDL and a ZV CDL and find that the former performs significantly better, meaning that sophisticatedly modeling uncertainty between two components is essential for performance.

\textbf{Learning Algorithms}: Due to the nature of BDL, practical learning algorithms need to meet the following criteria:
\begin{compactenum}
\item They should be online algorithms in order to scale well for large datasets.
\item They should be efficient enough to scale linearly with the number of free parameters in the perception component.
\end{compactenum}
Criterion (1) implies that conventional variational inference or MCMC methods are not applicable. Usually an online version of them is needed \cite{onlineVB}. Most SGD-based methods do not work either unless only MAP inference (as opposed to Bayesian treatments) is performed. Criterion (2) is needed because there are typically a large number of free parameters in the perception component. This means methods based on Laplace approximation \cite{mackay1992practical} are not realistic since they involve the computation of a Hessian matrix that scales quadratically with the number of free parameters.

%\begin{figure}[!tb]
%\begin{center}
%\includegraphics[height=2.0cm]{fig/BDL_typeII}
%\end{center}
%\vskip -0.2in
%\caption{PGM for type II BDL.
%}
%\label{fig:BDL_typeII}
%\vskip -0.1in
%\end{figure}
%
%\begin{figure}[!tb]
%\begin{center}
%\includegraphics[height=2.0cm]{fig/BDL_typeIII}
%\end{center}
%\vskip -0.2in
%\caption{PGM for type III BDL.
%}
%\label{fig:BDL_typeIII}
%\vskip -0.2in
%\end{figure}

\subsection{Perception Component}

Ideally, the perception component should be a probabilistic or Bayesian neural network, in order to be compatible with the task-specific component, which is probabilistic in nature. This is to ensure the perception component's built-in capability to handle uncertainty of parameters and its output.

As mentioned in Section~\ref{sec:history}, the study of Bayesian neural networks dates back to 1990s \cite{mackay1992practical,hinton1993keeping,neal1995bayesian,DBLP:conf/nips/Graves11}. However, pioneering work at that time was not widely adopted due to its lack of scalability. To address the this issue, there has been recent development such as restricted Boltzmann machine (RBM)~\cite{DBN-fast,RBM}, probabilistic generalized stacked denoising autoencoders (pSDAE)~\cite{RSDAE,CDL}, variational autoencoders (VAE)~\cite{kingma2013auto}, probabilistic back-propagation (PBP)~\cite{DBLP:conf/icml/Hernandez-Lobato15b}, Bayes by Backprop (BBB)~\cite{DBLP:conf/icml/BlundellCKW15}, Bayesian dark knowledge (BDK)~\cite{balan2015bayesian}, and natural-parameter networks (NPN)~\cite{NPN}.

More recently, generative adversarial networks (GAN)~\cite{GAN} prevail as a new training scheme for training neural networks and have shown promise in generating photo-realistic images. Later on, Bayesian formulations (as well as related theoretical results) for GAN have also been proposed~\cite{GAN,ProbGAN}. These models are also potential building blocks as the BDL framework's perception component.

In this subsection, we mainly focus on the introduction of recent Baysian neural networks such as RBM, pSDAE, VAE, and NPN. We refer the readers to \cite{dlbook} for earlier work in this direction.

% \textred{mention VAE and CVAE after CDL, echoing VAE}.

% \textred{cite thesis}

% \textred{cite back BDL survey}

% \textred{mention andrew's BDL report, cite some work}

% new development pSDAE VAE NPN

% difference between BNN and BDL, 

% BDL in a more general sense

\subsubsection{Restricted Boltzmann Machine}\label{sec:rbm}

Restricted Boltzmann Machine (RBM) is a special kind of BNN in that (1) it is not trained with back-propagation (BP) and that (2) its hidden neurons are binary. Specifically, RBM defines the following energy:

\begin{align*}
E(\v,\h)=-\v^T\W\h-\v^T\b-\h^T\a,
\end{align*}
where $\v$ denotes visible (observed) neurons, and $\h$ denotes binary hidden neurons. $\W$, $\a$, and $\b$ are learnable weights. The energy function leads to the following conditional distributions:
\begin{align}
p(\v | \h) = \frac{\exp(-E(\v,\h))}{\sum\limits_{\v}\exp(-E(\v,\h))},\;\;\;\;\;\;\;\;p(\h | \v) = \frac{\exp(-E(\v,\h))}{\sum\limits_{\h}\exp(-E(\v,\h))} \label{eq:rbm_conditional}
\end{align}

RBM is trained using `Contrastive Divergence'~\cite{DBN-fast} rather than BP. Once trained, RBM can infer $\v$ or $\h$ by marginalizing out other neurons. One can also stack layers of RBM to form a deep belief network (DBN)~\cite{DBN-speech}, use multiple branches of deep RBN for multimodal learning~\cite{MultRBM}, or combine DBN with convolutional layers to form a convolutional DBN~\cite{ConvDBN}. 

\subsubsection{Probabilistic Generalized SDAE}\label{sec:psdae}

Following the introduction of SDAE in Section \ref{sec:ae}, if we assume that both the clean input $\X_c$ and the corrupted input $\X_0$ are observed, similar to \cite{PRML,mackay1992practical,DBLP:conf/nips/BengioYAV13,nmSDAE},
%\footnote{*** They don't define a generative process, do they? (** References added and generative process modified.)}
we can define the following generative process of the probabilistic SDAE:
\begin{enumerate}
\item For each layer $l$ of the SDAE network,
\begin{enumerate}
\item For each column $n$ of the weight matrix $\W_l$, draw 
% \begin{align*}
$
\W_{l,*n} \sim \NM(\0,\lambda_w^{-1} \I_{K_l}).
$
% \end{align*}
\item Draw the bias vector $\b_l \sim \NM(\0,\lambda_w^{-1} \I_{K_l})$.
\item For each row $j$ of $\X_l$, draw
\begin{align}\label{eq:gaussian}
\X_{l,j*} \sim \NM(\sigma(\X_{l-1,j*}\W_l+\b_l),\lambda_s^{-1} \I_{K_l}).
%\X_{l,j*} \sim \mbox{Delta}(\sigma(\X_{l-1,j*}\W_l+\b_l)).
\end{align}
%\item Generate the output of layer $l$, $\X_l = \sigma(\X_{l-1}\W_l+\b_l)$.
\end{enumerate}
\item For each item $j$, draw a clean input\footnote{Note that while generation of the \emph{clean} input $\X_c$ from $\X_L$ is part of the generative process of the Bayesian SDAE, generation of the \emph{noise-corrupted} input $\X_0$ from $\X_c$ is an artificial noise injection process to help the SDAE learn a more robust feature representation.} 
% \begin{align*}
$
\X_{c,j*} \sim \NM(\X_{L,j*},\lambda_n^{-1}\I_{B}).
$
% \end{align*}
%\footnote{*** So you assume that it has unit variance? (** Yes.) (*** Why isn't $\lambda_n$ introduced here, but later?) (** It might be better to introduce it here to keep it consistent with the latter. Delta distribution is used so that $\X$ can be treated as `generalized' random variables drawn from some distribution.)}
\end{enumerate}
Note that if $\lambda_s$ goes to infinity, the Gaussian distribution in Equation (\ref{eq:gaussian}) will become a Dirac delta distribution \cite{strichartz2003guide} centered at $\sigma(\X_{l-1,j*}\W_l+\b_l)$, where $\sigma(\cdot)$ is the sigmoid function, and the model will degenerate into a Bayesian formulation of vanilla SDAE. This is why we call it `generalized' SDAE.
%We note that it is equivalent to a Gaussian distribution
%\begin{align*}
%\NM(\sigma(\X_{l-1,j*}\W_l+\b_l),\lambda^{-1} \I_{K_l})
%\end{align*}
%with $\lambda$ approaching infinity.

%\footnote{** Dirac delta distribution is the core of real analysis and is one of the most important building block in signal processing and physics. That means there is a deep background for using Dirac delta function. Should we mention this in the paper? (*** I don't think you have space for it.)}

%where $J$ is the number of items, $L$ is the number of layers, $\W_{l,*n}$ means the n'th column of weight matrix $\W_l$, and $\X_{c,j*}$ is the j'th row of the clean input matrix $\X_c$.
%\footnote{*** Is it necessary to mention this again as you have already introduced it in Section 2? (** Deleted redundant part, and moved non-redundant part to Section 2.)}

The first $L/2$ layers of the network act as an encoder and the last $L/2$ layers act as a decoder.  Maximization of the posterior probability is equivalent to minimization of the reconstruction error with weight decay taken into consideration.
%Note that practice the weight decay of $\b_l$ is $0$, which makes the step 1(c) unnecessary.
%\footnote{*** This sentence may not be very clear to readers. (** Deleted. Actually, in the my code deepmat and deeplearntoolbox, weight decay are also used for biases, which seem to work well. To avoid confusion, I guess we'd better delete this sentence and use weigh decay also for biases in our models.)}

Following pSDAE, both its convolutional version~\cite{CKE} and its recurrent version~\cite{CRAE} have been proposed with applications in knowledge base embedding and recommender systems.

\subsubsection{Variational Autoencoders}\label{sec:vae}

Variational Autoencoders (VAE)~\cite{kingma2013auto} essentially tries to learn parameters $\phi$ and $\theta$ that maximize the evidence lower bound (ELBO):
\begin{align}
\mathcal{L}_{vae} = E_{q_{\phi}(\z | \x)}[\log p_{\theta}(\x | \z)] - KL(q_{\phi}(\z | \x) \| p(\z)), \label{eq:vae_elbo}
\end{align}
where $q_{\phi}(\z | \x)$ is the encoder parameterized by $\phi$ and $p_{\theta}(\x | \z)$ is the decoder parameterized by $\theta$. The negation of the first term is similar to the reconstrunction error in vanilla AE, while the KL divergence works as a regularization term for the encoder. During training $q_{\phi}(\z | \x)$ will output the mean and variance of a Gaussian distribution, from which $\z$ is sampled via the reparameterization trick. Usually $q_{\phi}(\z | \x)$ is parameterized by an MLP with two branches, one producing the mean and the other producing the variance. 

Similar to the case of pSDAE, various VAE variants have been proposed. For example, Importance weighted Autoencoders (IWAE)~\cite{IWAE} derived a tighter lower bound via importance weighting, \cite{LSTM-VAE-CNN} combined LSTM, VAE, and dilated CNN for text modeling, \cite{VRNN} proposed a recurrent version of VAE dubbed variational RNN (VRNN).

\subsubsection{Natural-Parameter Networks}
Different from vanilla NN which usually takes deterministic input, NPN~\cite{NPN} is a probabilistic NN taking distributions as input. The input distributions go through layers of linear and nonlinear transformation to produce output distributions. In NPN, all hidden neurons and weights are also distributions expressed in closed form. Note that this is in contrast to VAE where only the middle layer output $\z$ is a distribution. 

As a simple example, in a vanilla linear NN $f_w(x)=wx$ takes a scalar $x$ as input and computes the output based on a scalar parameter $w$; a corresponding Gaussian NPN would assume $w$ is drawn from a Gaussian distribution $\mathcal{N}(w_m, w_s)$ and that $x$ is drawn from $\mathcal{N}(x_m, x_s)$ ($x_s$ is set to $0$ when the input is deterministic). With $\theta=(w_m,w_s)$ as a learnable parameter pair, NPN will then compute the mean and variance of the output Gaussian distribution $\mu_{\theta}(x_m, x_s)$ and $s_{\theta}(x_m, x_s)$ in closed form (bias terms are ignored for clarity) as:
\begin{align}
\mu_{\theta}(x_m,x_s)&=E[wx]=x_m w_m, \label{eq:npn_mean} \\
s_{\theta}(x_m,x_s)&=D[wx]=x_s w_s+x_s w_m^2+x_m^2 w_s, \label{eq:npn_var}
\end{align}
Hence the output of this Gaussian NPN is a tuple $(\mu_{\theta}(x_m, x_s),s_{\theta}(x_m, x_s))$ representing a Gaussian distribution instead of a single value. Input variance $x_s$ to NPN can be set to $0$ if not available. Note that since $s_{\theta}(x_m,0)=x_m^2 w_s$, $w_m$ and $w_s$ can still be learned even if $x_s=0$ for all data points. The derivation above is generalized to handle vectors and matrices in practice~\cite{NPN}. Besides Gaussian distributions, NPN also support other exponential-family distributions such as Poisson distributions and gamma distributions~\cite{NPN}.
%The output of this NPN, $\mu_{\theta}(x_m, x_s)$ and $s_{\theta}(x_m, x_s)$, are the mean and variance of the following distribution:
%\begin{align*}
%p(o|x_m,x_s,w_m,w_s) = E_{p(x|x_m,x_s) p(w|w_m,w_s)}[ p(o|x,w)],
%\end{align*}
%where $p(o|x,w)$ is a Dirac delta distribution centered at $xw$ (see the Supplement for generalization of NPN to vectors and matrices). These properties turn out to be the key to efficient computation of marginal likelihood (details below).

Following NPN, a light-weight version~\cite{LightNPN} was proposed to speed up the training and inference process. Another variant, MaxNPN~\cite{MaxNPN}, extended NPN to handle max-pooling and categorical layers. ConvNPN~\cite{ConvNPN} enables convolutional layers in NPN. In terms of model quantization and compression, BinaryNPN~\cite{BinaryNPN} was also proposed as NPN's binary version to achieve better efficiency.

% \textred{subsubsection RBM? mention similar RBM follow-ups?}
% \textred{need to rewrite sentences to avoid overlapping with BIN paper}

\subsection{Task-Specific Component}
In this subsection, we introduce different forms of task-specific components. The purpose of a task-specific component is to incorporate probabilistic prior knowledge into the BDL model. Such knowledge can be naturally represented using PGM. Concretely, it can be a typical (or shallow) Bayesian network~\cite{BayesNetBook,PRML}, a bidirectional inference network~\cite{BIN}, or a stochastic process~\cite{ross1996stochastic}. 

\subsubsection{Bayesian Networks}\label{sec:bayesnet}
Bayesian networks are the most common choice for a task-specific component. As mentioned in Section~\ref{sec:pgm}, Bayesian networks can naturally represent conditional dependencies and handle uncertainty. Besides LDA introduced above, a more straightforward example is probabilistic matrix factorization (PMF)~\cite{PMF}, where one uses a Bayesian network to describe the conditional dependencies among users, items, and ratings. Specifically, PMF assumes the following generative process:

\begin{compactenum}
\item For each item $j$, draw a latent item vector:
$
\v_i \sim \NM(\0,\lambda_v^{-1}\I_K).
$
\item For each user $i$, draw a latent user vector:
$
\u_i \sim \NM(\0,\lambda_u^{-1}\I_K).
$
\item For each user-item pair $(i,j)$, draw a rating:
$
\R_{ij} \sim \NM(\u_i^T\v_j,\C_{ij}^{-1}).
$
\end{compactenum}

In the generative process above, $\C_{ij}^{-1}$ is the corresponding variance for the rating $\R_{ij}$. Using MAP estimates, learning PMF amounts to maximize the following log-likelihood of $p(\{\u_i\}, \{\v_j\} | \{\R_{ij}\}, \{\C_{ij}\}, \lambda_u, \lambda_v)$:
\begin{align*}
\mathscr{L}=-\frac{\lambda_u}{2}\sum\limits_i \|\u_i\|_2^2
-\frac{\lambda_v}{2}\sum\limits_j \|\v_j\|_2^2 
-\sum\limits_{i,j}\frac{\C_{ij}}{2}(\R_{ij}-\u_i^T\v_j)^2,
\end{align*}

Note that one can also impose another layer of priors on the hyperparameters with a fully Bayesian treatment. For example, \cite{BPMF} imposes priors on the precision matrix of latent factors and learn the Bayesian PMF with Gibbs sampling. 

In Section~\ref{sec:recsys}, we will show how PMF can be used as a task-specific component along with a perception component defined to significantly improve recommender systems' performance. 

% PMF CDL

\subsubsection{Bidirectional Inference Networks}\label{sec:bin}
Typical Bayesian networks assume `shallow' conditional dependencies among random variables. In the generative process, one random variable (which can be either latent or observed) is usually drawn from a conditional distribution parameterized by the linear combination of its parent variables. For example, in PMF the rating $\R_{ij}$ is drawn from a Gaussian distribution mainly parameterized by the linear combination of $\u_i$ and $\v_j$, i.e., $\R_{ij} \sim \NM(\u_i^T\v_j,\C_{ij}^{-1})$. 

\begin{figure}[!tb]
\begin{center}
%\framebox[4.0in]{$\;$}
%\includegraphics[height=5cm]{likeli1.eps}
\subfigure{
\includegraphics[height=2.6cm]{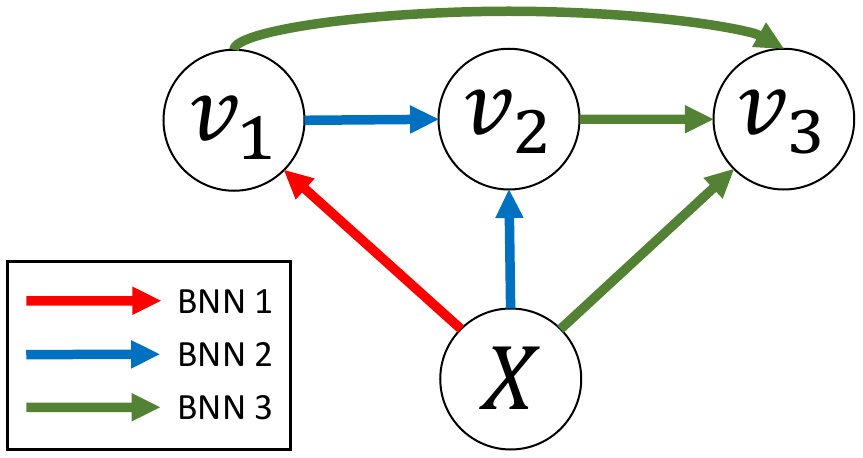}}
\hspace{0.5in}
\subfigure{
\includegraphics[height=2.6cm]{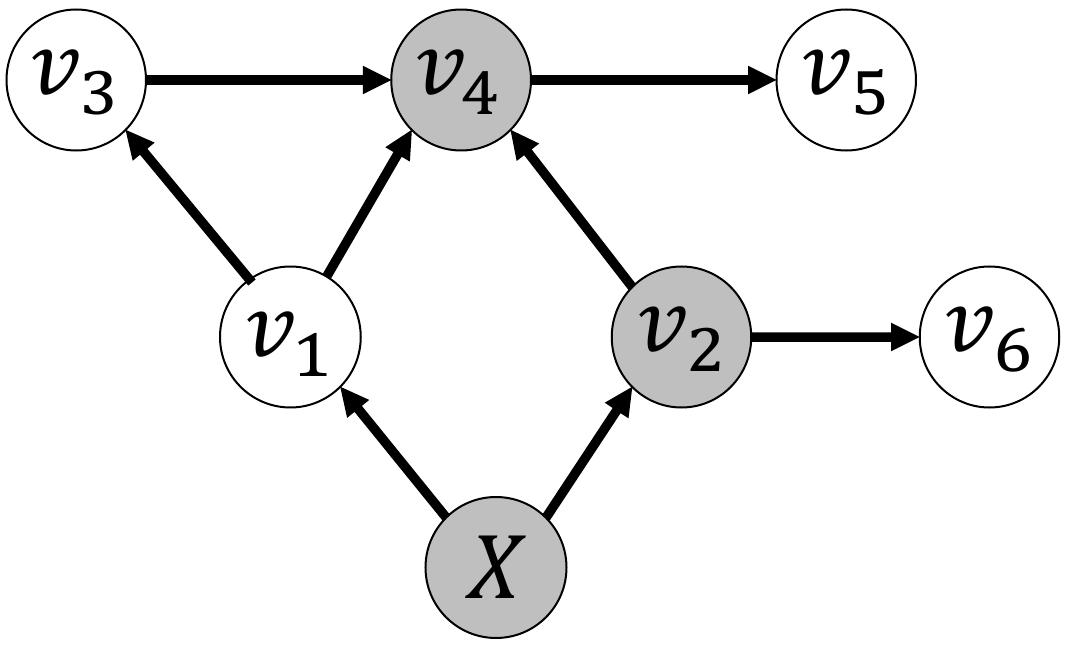}}
\end{center}
\vskip -0.2in
\caption{Left: A simple example of BIN with each conditional distribution parameterized by a Bayesian neural networks (BNN) or simply a probabilistic neural network. Right: Another example BIN. Shaded and transparent nodes indicate observed and unobserved variables, respectively.
}
\label{fig:bin}
\vskip -0.2in
\end{figure}

Such `shallow' and linear structures can be replaced with nonlinear or even deep nonlinear structures to form a \emph{deep Bayesian network}. As an example, bidirectional inference network (BIN)~\cite{BIN} is a class of deep Bayesian networks that enable deep nonlinear structures in each conditional distribution, while retaining the ability to incorporate prior knowledge as Bayesian networks.

For example, Figure~\ref{fig:bin}(left) shows a BIN, where each conditional distribution is parameterized by a Bayesian neural network. Specifically, this example assumes the following factorization:
\begin{align*}
p(v_1, v_2, v_3 | X) = p(v_1 | X) p(v_2 | X, v_1) p(v_3 | X, v_1, v_2).
\end{align*}
A vanilla Bayesian network parameterizes each distribution with simple linear operations. For example, $p(v_2 | X, v_1) = \NM (v_2| Xw_0 + v_1 w_1 + b, \sigma^2)$). In contrast, BIN (as a deep Bayesian network) uses a BNN. For example, BIN has $p(v_2 | X, v_1) = \NM (v_2| \mu_{\theta}(X, v_1), s_{\theta}(X, v_1))$, where $\mu_{\theta}(X, v_1)$ and $s_{\theta}(X, v_1)$ are the output mean and variance of the BNN. The inference and learning of such a deep Bayesian network is done by performing BP across all BNNs (e.g., BNN 1, 2, and 3 in Figure~\ref{fig:bin}(left))~\cite{BIN}.

Compared to vanilla (shallow) Bayesian networks, deep Bayesian networks such as BIN make it possible to handle deep and nonlinear conditional dependencies effectively and efficiently. Besides, with BNN as building blocks, task-specific components based on deep Bayesian networks can better work with the perception component which is usually a BNN as well. Figure~\ref{fig:bin}(right) shows a more complicated case with both observed (shaded nodes) and unobserved (transparent nodes) variables. 

\subsubsection{Stochastic Processes}
Besides vanilla Bayesian networks and deep Bayesian networks, a task-specific component can also take the form of a stochastic process~\cite{ross1996stochastic}. For example, a Wiener process can naturally describe a continuous-time Brownian motion model $\x_{t+u} | \x_t \sim \NM(\x_t, \lambda u \I$), where $\x_{t+u}$ and $\x_t$ are the states at time $t$ and $t+u$, respectively. In the graphical model literature, such a process has been used to model the continuous-time topic evolution of articles over time~\cite{cDTM}. 

Another example is to model phonemes' boundary positions using a Poisson process in automatic speech recognition (ASR)~\cite{RPPU}. Note that this is a fundamental problem in ASR since speech is no more than a sequence of phonemes. Specifically, a Poisson process defines the generative process $\Delta t_i = t_i - t_{i-1} \sim g(\lambda(t))$, with $\mathcal{T} = \{t_1, t_2, \dots, t_N\}$ as the set of boundary positions, and $g(\lambda(t))$ is a exponential distribution with the parameter $\lambda(t)$ (also known as the intensity). Such a stochastic process naturally models the occurrence of phoneme boundaries in continuous time. The parameter $\lambda(t)$ can be the output of a neural network taking raw speech signals as input~\cite{RPPU,IFL,NN-SP}. 

Interestingly, stochastic processes can be seen as a type of dynamic Bayesian networks. To see this, we can rewrite the Poisson process above in an equivalent form, where given $t_{i-1}$, the probability that $t_i$ has not occurred at time $t$, $P(t_i > t) = \exp(\int_{t_{i-1}}^t -\lambda(t) d t)$. Obviously both the Wiener process and the Poisson process are Markovian and can be represented with a dynamic Bayesian network~\cite{murphybook}. 

For clarity, we focus on using vanilla Bayesian networks as task-specific components in Section~\ref{sec:app}; they can be naturally replaced with other types of task-specific components to represent different prior knowledge if necessary.

% phoneme boundary, inter-arrival time

% Poisson process \cite{RPPU}
% Wiener process continuous-time dynamic topic model \cite{cDTM}
% \cite{RPPU,IFL,NN-SP}

\section{Concrete BDL Models and Applications}\label{sec:app}
In this section, we discuss how the BDL framework can facilitate supervised learning, unsupervised learning, and representation learning in general. Concretely we use examples in domains such as recommender systems, topic models, control, etc. 

% concrete models under the BDL framework and their applications on supervised learning, unsupervised learning, and representation learning in general. Concretely we various domains such as recommender systems, topic models, control, etc.

\subsection{Supervised Bayesian Deep Learning for Recommender Systems}\label{sec:recsys}
Despite the successful applications of deep learning on natural language processing and computer vision, very few attempts have been made to develop deep learning models for collaborative filtering (CF) before the emergence of BDL. \cite{DBLP:conf/icml/SalakhutdinovMH07} uses restricted Boltzmann machines instead of the conventional matrix factorization formulation to perform CF and \cite{DBLP:conf/icml/GeorgievN13} extends this work by incorporating user-user and item-item correlations. Although these methods involve both deep learning and CF, they actually belong to CF-based methods because they ignore users' or items' content information, which is crucial for accurate recommendation. \cite{DBLP:conf/icassp/SainathKSAR13} uses low-rank matrix factorization in the last weight layer of a deep network to significantly reduce the number of model parameters and speed up training, but it is for classification instead of recommendation tasks. On music recommendation, \cite{DBLP:conf/nips/OordDS13,DBLP:conf/mm/WangW14} directly use conventional CNN or deep belief networks (\mbox{DBN}) to assist representation learning for content information, but the deep learning components of their models are deterministic without modeling the noise and hence they are less robust. The models achieve performance boost mainly by loosely coupled methods without exploiting the interaction between content information and ratings. Besides, the CNN is linked directly to the rating matrix, which means the models will perform poorly due to serious overfitting when the ratings are sparse.

\subsubsection{Collaborative Deep Learning}
To address the challenges above, a hierarchical Bayesian model called collaborative deep learning (CDL) as a novel tightly coupled method for recommender systems is introduced in \cite{CDL}. Based on a Bayesian formulation of SDAE, CDL tightly couples deep representation learning for the content information and collaborative filtering for the rating (feedback) matrix, allowing two-way interaction between the two. From BDL's perspective, a probabilistic SDAE as the perception component is tightly coupled with a probabilistic graphical model as the task-specific component. Experiments show that CDL significantly improves upon the state of the art.

In the following text, we will start with the introduction of the notation used during our presentation of CDL. After that we will review the design and learning of CDL.

\textbf{Notation and Problem Formulation}:
Similar to the work in \cite{CTR}, the recommendation task considered in CDL takes implicit feedback \cite{DBLP:conf/icdm/HuKV08} as the training and test data.  The entire collection of $J$ items (articles or movies) is represented by a $J$-by-$B$ matrix $\X_c$, where row $j$ is the bag-of-words vector $\X_{c,j*}$ for item $j$ based on a vocabulary of size $B$.  With $I$ users, we define an $I$-by-$J$ binary rating matrix $\R=[\R_{ij}]_{I\times J}$. For example, in the dataset \emph{citeulike-a} \cite{CTR,CTRSR,CDL} $\R_{ij}=1$ if user $i$ has article $j$ in his or her personal library and $\R_{ij}=0$ otherwise. Given part of the ratings in $\R$ and the content information $\X_c$, the problem is to predict the other ratings in $\R$. Note that although CDL in its current from focuses on movie recommendation (where plots of movies are considered as content information) and article recommendation like \cite{CTR} in this section, it is general enough to handle other recommendation tasks (e.g., tag recommendation).

The matrix $\X_c$ plays the role of clean input to the SDAE while the noise-corrupted matrix, also a $J$-by-$B$ matrix, is denoted by $\X_0$.  The output of layer $l$ of the SDAE is denoted by $\X_l$ which is a $J$-by-$K_l$ matrix.  Similar to $\X_c$, row $j$ of $\X_l$ is denoted by $\X_{l,j*}$.  $\W_l$ and $\b_l$ are the weight matrix and bias vector, respectively, of layer $l$, $\W_{l,*n}$ denotes column $n$ of $\W_l$, and $L$ is the number of layers. For convenience, we use $\W^+$ to denote the collection of all layers of weight matrices and biases. Note that an $L/2$-layer SDAE corresponds to an $L$-layer network.

\begin{figure*}[!tb]
\begin{center}
\subfigure{
\includegraphics[height=3.6cm]{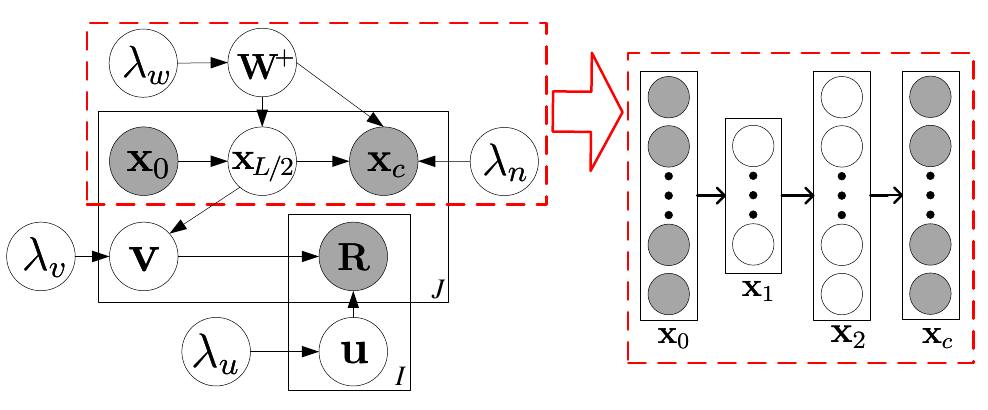}}
\hspace{0.3in}
\subfigure{
\includegraphics[height=3.4cm]{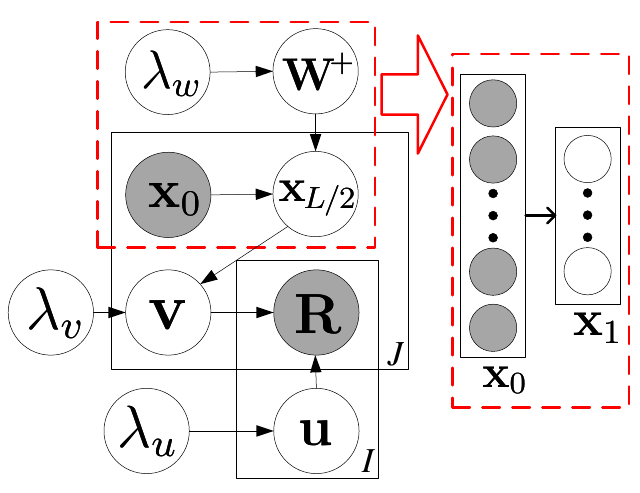}}
\end{center}
\vskip -0.2in
\caption{On the left is the graphical model of CDL. The part inside the dashed rectangle represents an SDAE.  An example SDAE with $L = 2$ is shown. On the right is the graphical model of the degenerated CDL. The part inside the dashed rectangle represents the encoder of an SDAE. An example SDAE with $L=2$ is shown on its right. Note that although $L$ is still $2$, the decoder of the SDAE vanishes. To prevent clutter, we omit all variables $\x_l$ except $\x_0$ and $\x_{L/2}$ in the graphical models.
}
\label{fig:cdl_pgm}
\vskip -0.2in
\end{figure*}

\textbf{Collaborative Deep Learning}: Using the probabilistic SDAE in Section \ref{sec:psdae} as a component, the generative process of CDL is defined as follows:\\
\begin{compactenum}
\item For each layer $l$ of the SDAE network,
\begin{compactenum}
\item For each column $n$ of the weight matrix $\W_l$, draw
$\W_{l,*n} \sim \NM(\0,\lambda_w^{-1} \I_{K_l})$.
\item Draw the bias vector $\b_l \sim \NM(\0,\lambda_w^{-1} \I_{K_l})$.
\item For each row $j$ of $\X_l$, draw 
$
\X_{l,j*} \sim \NM(\sigma(\X_{l-1,j*}\W_l+\b_l),\lambda_s^{-1} \I_{K_l}).
$
\end{compactenum}
\item For each item $j$,
\begin{compactenum}
\item Draw a clean input $\X_{c,j*} \sim \NM(\X_{L,j*},\lambda_n^{-1} \I_{J}$).
\item Draw the latent item offset vector $\ep_j \sim \NM(\0,\lambda_v^{-1}\I_K)$ and then set the latent item vector:
$
\v_j=\ep_j+\X_{\frac{L}{2},j*}^T.
$
\end{compactenum}
\item Draw a latent user vector for each user $i$: 
$
\u_i \sim \NM(\0,\lambda_u^{-1}\I_K).
$
\item Draw a rating $\R_{ij}$ for each user-item pair $(i,j)$:
$
\R_{ij} \sim \NM(\u_i^T\v_j,\C_{ij}^{-1}).\\
$
\end{compactenum}

Here $\lambda_w$, $\lambda_n$, $\lambda_u$, $\lambda_s$, and $\lambda_v$ are hyperparameters
%\footnote{*** Why is $\lambda_n$ mentioned specifically but not the other hyperparameters? (** Modified.)}
and $\C_{ij}$ is a confidence parameter similar to that for CTR \cite{CTR} ($\C_{ij} = a$ if $\R_{ij}=1$ and $\C_{ij}=b$ otherwise). Note that the middle layer $\X_{L/2}$ serves as a bridge between the ratings and content information. This middle layer, along with the latent offset $\ep_j$, is the key that enables CDL to simultaneously learn an effective feature representation and capture the similarity and (implicit) relationship among items (and users). Similar to the generalized SDAE, we can also take $\lambda_s$ to infinity for computational efficiency.

The graphical model of CDL when $\lambda_s$ approaches positive infinity is shown in Figure \ref{fig:cdl_pgm}, where, for notational simplicity, we use $\x_0$, $\x_{L/2}$, and $\x_L$ in place of $\X_{0,j*}^T$, $\X_{\frac{L}{2},j*}^T$, and $\X_{L,j*}^T$, respectively.

Note that according the definition in Section \ref{sec:general}, here the perception variables $\Om_p=\{\{\W_l\},\{\b_l\},\{\X_l\},\X_c\}$, the hinge variables $\Om_h=\{\V\}$, and the task variables $\Om_t=\{\U,\R\}$.
%$\W^+$ is a shorthand for weights $\W_l$ and biases $\b_l$ for all $L$ layers.
%\footnote{*** Already introduced in Section 2. (** Deleted.)}
%\footnote{*** The last two sentences are redundant.  Either keep them here or in the figure capture, but not both. (** Deleted.)}

%\begin{figure} [tb]
%\begin{center}
%  \begin{tabular}{ccc}
%   \includegraphics*[height=35mm]{fig/cdl_with_sdae}
%  \end{tabular} %\vskip -0.4cm
%\caption{\small Graphical model of CDL. The part inside the dashed rectangle represents an SDAE.  A sample SDAE with $L = 2$ is shown on the right-hand side.}
%\label{fig:cdl_graphmodel}
%\end{center}
%\vskip -0.5cm
%\end{figure}
\textbf{Learning}:
Based on the CDL model above, all parameters could be treated as random variables so that fully Bayesian methods such as Markov chain Monte Carlo (MCMC) or variational inference \cite{DBLP:journals/ml/JordanGJS99} may be applied.  However, such treatment typically incurs high computational cost. Therefore CDL uses an EM-style algorithm to obtain the MAP estimates, as in \cite{CTR}.

Concretely, note that maximizing the posterior probability is equivalent to maximizing the joint log-likelihood of $\U$, $\V$, $\{\X_l\}$, $\X_c$, $\{\W_l\}$, $\{\b_l\}$, and $\R$ given $\lambda_u$, $\lambda_v$, $\lambda_w$, $\lambda_s$, and $\lambda_n$:
\begin{align}
\mathscr{L}=&-\frac{\lambda_u}{2}\sum\limits_i \|\u_i\|_2^2
-\frac{\lambda_w}{2}\sum\limits_l(\|\W_l\|_F^2+\|\b_l\|_2^2) -\frac{\lambda_v}{2}\sum\limits_j\|\v_j-\X_{\frac{L}{2},j*}^T\|_2^2-\frac{\lambda_n}{2}\sum\limits_j\|\X_{L,j*}-\X_{c,j*}\|_2^2 \nonumber\\ &-\frac{\lambda_s}{2}\sum\limits_l\sum\limits_j\|\sigma(\X_{l-1,j*}\W_l+\b_l)-\X_{l,j*}\|_2^2 -\sum\limits_{i,j}\frac{\C_{ij}}{2}(\R_{ij}-\u_i^T\v_j)^2. \nonumber
\end{align}\label{eq:Lgen}
If $\lambda_s$ goes to infinity, the likelihood becomes:
%\footnote{*** Better use $\{\X_l\}, \{\W_l\}, \{\b_l\}$. (** Modified. I also added the indexes but don't know if it's necessary. Please delete them if it is not since it looks kind of messy with the indexes.)}
%\begin{align}\label{eq:L}
%\mathscr{L}=&-\frac{\lambda_u}{2}\sum\limits_i \u_i^T\u_i
%-\frac{\lambda_w}{2}\sum\limits_l(\|\W_l\|_F^2+\|\b_l\|_2^2) \nonumber \\
%&-\frac{\lambda_v}{2}\sum\limits_j(\v_j-f_e(\X_{0,j*},\W^+))^T(\v_j-f_e(\X_{0,j*},\W^+)) \nonumber \\
%&-\frac{\lambda_n}{2}\sum\limits_j(f_r(\X_{0,j*},\W^+)-\X_{c,j*})^T(f_r(\X_{0,j*},\W^+)-\X_{c,j*})
%-\sum\limits_{i,j}\frac{c_{ij}}{2}(r_{ij}-\u_i^T\v_j)^2,
%\nonumber
%\end{align}
\begin{align}\label{eq:L}
\mathscr{L}=&-\frac{\lambda_u}{2}\sum\limits_i \|\u_i\|_2^2
-\frac{\lambda_w}{2}\sum\limits_l(\|\W_l\|_F^2+\|\b_l\|_2^2)\nonumber -\frac{\lambda_v}{2}\sum\limits_j\|\v_j-f_e(\X_{0,j*},\W^+)^T\|_2^2 \nonumber \\
&-\frac{\lambda_n}{2}\sum\limits_j\|f_r(\X_{0,j*},\W^+)-\X_{c,j*}\|_2^2 -\sum\limits_{i,j}\frac{\C_{ij}}{2}(\R_{ij}-\u_i^T\v_j)^2,
\end{align}
where the encoder function $f_e(\cdot,\W^+)$ takes the corrupted content vector $\X_{0,j*}$ of item $j$ as input and computes its encoding, and the function $f_r(\cdot,\W^+)$ also takes $\X_{0,j*}$ as input, computes the encoding and then reconstructs item $j$'s content vector.
%\footnote{*** The name reconstruction function is not exactly correct because it actually performs \emph{both} encoding and reconstruction, taking $\X_{0,j*}$ as input. (** Modified. But don't know if it is clear enough.)}
For example, if the number of layers $L=6$, $f_e(\X_{0,j*},\W^+)$ is the output of the third layer while $f_r(\X_{0,j*},\W^+)$ is the output of the sixth layer.

\begin{figure}[!tb]
\begin{center}
%\framebox[4.0in]{$\;$}
%\includegraphics[height=5cm]{likeli1.eps}
\vskip -0.23in
\subfigure{
\includegraphics[height=5.5cm]{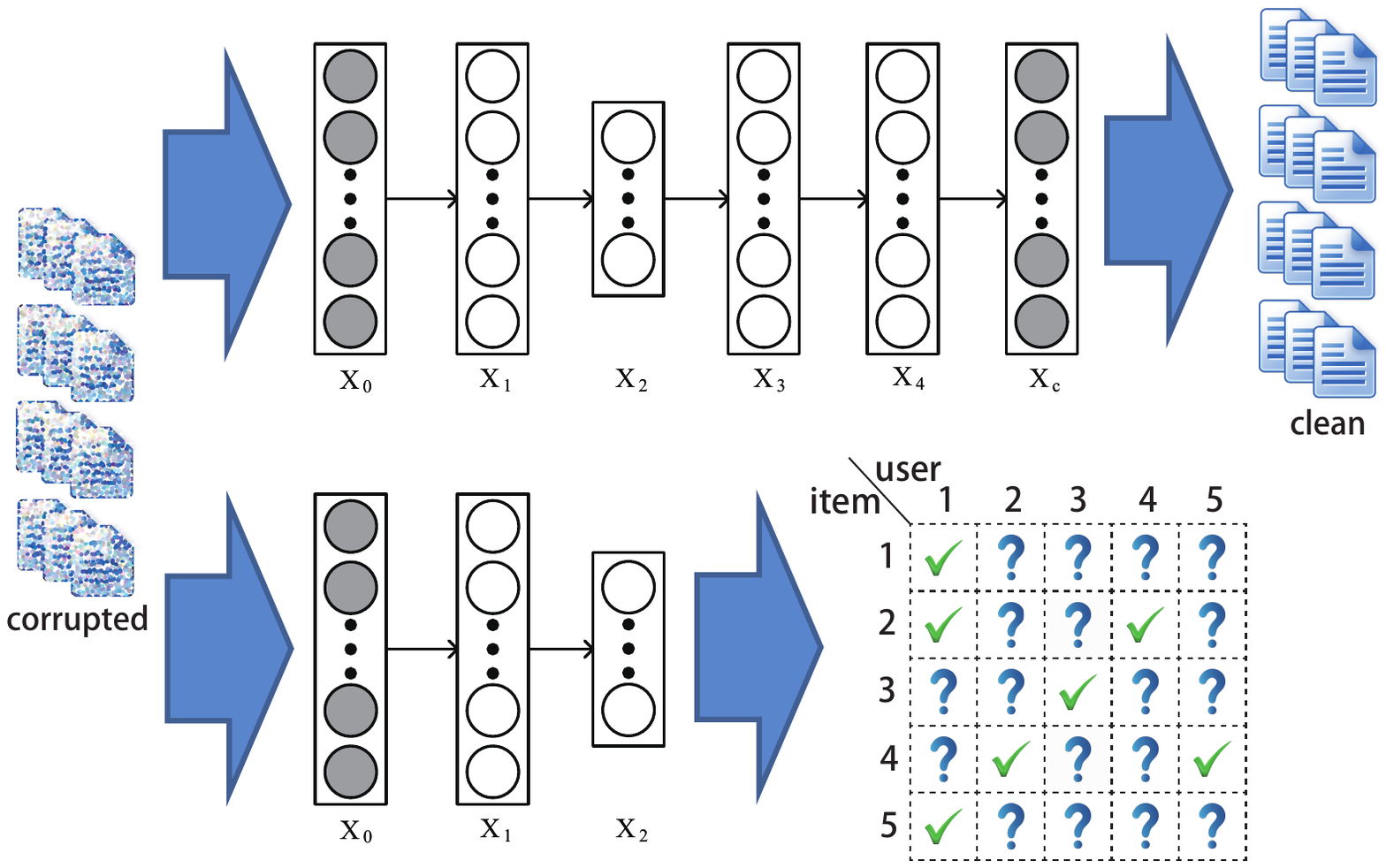}}%5.0
\hspace{0.0in}%0.3
\subfigure{
\includegraphics[height=5.1cm]{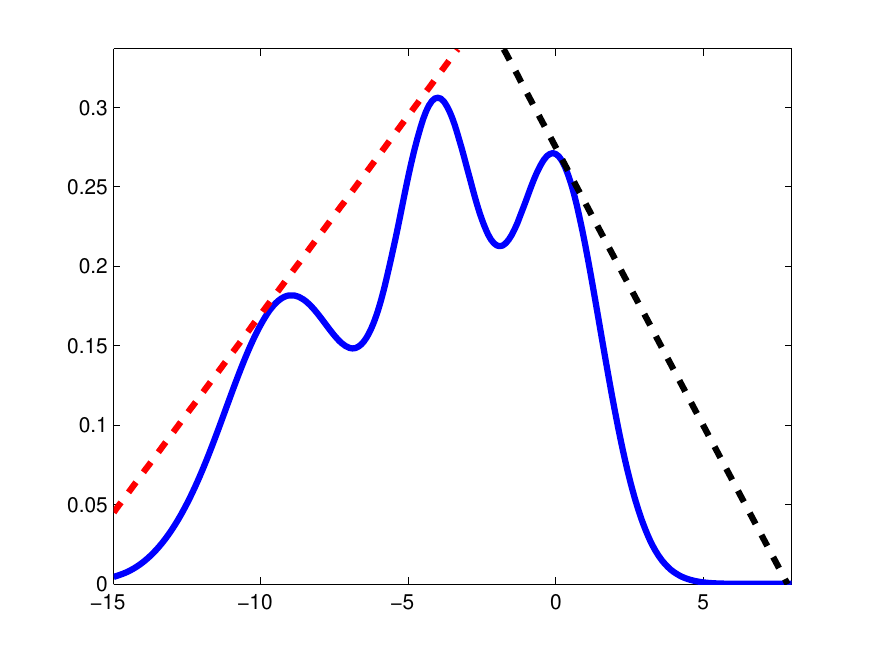}}%4.6
\end{center}
\vskip -0.3in
\caption{Left: NN representation for degenerated CDL. Right: Sampling as generalized BP in Bayesian CDL. 
}
\label{fig:twonet}
\vskip -0.6in
\end{figure}

%\begin{align*}
%\mathscr{L}=&-\frac{\lambda_u}{2}\sum\limits_i \|\u_i\|_2^2
%-\frac{\lambda_v}{2}\sum\limits_j\|\v_j-f_e(\X_{0,j*},\W^+)^T\|_2^2  \\
%&-\frac{\lambda_n}{2}\sum\limits_j\|f_r(\X_{0,j*},\W^+)-\X_{c,j*}\|_2^2 \\
%&-\sum\limits_{i,j}\frac{\C_{ij}}{2}(\R_{ij}-\u_i^T\v_j)^2,
%\end{align*}

From the perspective of optimization, the third term in the objective function, i.e., Equation (\ref{eq:L}), above is equivalent to a multi-layer perceptron using the latent item vectors $\v_j$ as the target while the fourth term is equivalent to an SDAE minimizing the reconstruction error. Seeing from the view of neural networks (NN), when $\lambda_s$ approaches positive infinity, training of the probabilistic graphical model of CDL in Figure \ref{fig:cdl_pgm}(left) would degenerate to simultaneously training two neural networks overlaid together with a common input layer (the corrupted input) but different output layers, as shown in Figure \ref{fig:twonet}(left). Note that the second network is much more complex than typical neural networks due to the involvement of the rating matrix.
%\footnote{Note that empirically the bias $\b_l$ does not need to have regularizer (corresponding to weight decay in neural network training). Since in our experiments the existence of weight decay term for bias $\b_l$ has little impact on the performance, we will keep the term here.}
%\footnote{*** I find the remark in the previous footnote is a bit strange.  Since weight decay for the bias has little impact, why is it still kept in the equation? (** Footnote deleted. Since in the code we also use weight decay on biases, which works just as well. I guess we will skip the footnote to avoid confusion.)}

When the ratio $\lambda_n/\lambda_v$ approaches positive infinity, it will degenerate to a two-step model in which the latent representation learned using SDAE is put directly into the CTR.  Another extreme happens when $\lambda_n/\lambda_v$ goes to zero where the decoder of the SDAE essentially vanishes.  Figure \ref{fig:cdl_pgm}(right) shows the graphical model of the degenerated CDL when $\lambda_n/\lambda_v$ goes to zero.
%\footnote{*** Again these two sentences are redundant. (** Deleted.)}
As demonstrated in experiments, the predictive performance will suffer greatly for both extreme cases \cite{CDL}.

%\begin{figure} [tb]
%\begin{center}
%  \begin{tabular}{ccc}
%   \includegraphics*[height=35mm]{fig/degenerated_cdl_with_sdae}
%  \end{tabular} %\vskip -0.4cm
%\caption{\small Graphical model of degenerated CDL. The part inside the dashed rectangle represents the encoder of an SDAE. A sample SDAE with $L=2$ is shown on the right-hand side.  Note that although $L$ is still $2$, the decoder of the SDAE vanishes.} \label{fig:pgm_cdl_encode}
%\end{center}
%\end{figure}
For $\u_i$ and $\v_j$, block coordinate descent similar to \cite{CTR,DBLP:conf/icdm/HuKV08} is used. Given the current $\W^+$, we compute the gradients of $\mathscr{L}$ with respect to $\u_i$ and $\v_j$ and then set them to zero, leading to the following update rules:
\begin{align}
\u_i&\leftarrow(\V \C_i \V^T+\lambda_u \I_K)^{-1}\V \C_i \R_i, \nonumber \;\;\;\;
\v_j\leftarrow(\U \C_i \U^T+\lambda_v \I_K)^{-1}(\U \C_j \R_j+\lambda_v f_e(\X_{0,j*},\W^+)^T), \nonumber
\end{align}
where $\U=(\u_i)^I_{i=1}$, $\V=(\v_j)^J_{j=1}$, $\C_i = \mbox{diag}(\C_{i1},\ldots,\C_{iJ})$ is a diagonal matrix,
%\footnote{*** $\C_i = \mbox{diag}(c_{i1},\ldots,c_{iJ})$ (** Modified.)}
$\R_i = (\R_{i1},\ldots,\R_{iJ})^T$ is a column vector containing all the ratings of user $i$,
%\footnote{*** $\R_i = (r_{i1},\ldots,r_{iJ})^T$ (** Modified.)}
and $\C_{ij}$ reflects the confidence controlled by $a$ and $b$ as discussed in \cite{DBLP:conf/icdm/HuKV08}. $\C_j$ and $\R_j$ are defined similarly for item $j$.

Given $\U$ and $\V$, we can learn the weights $\W_l$ and biases $\b_l$ for each layer using the back-propagation learning algorithm. The gradients of the likelihood with respect to $\W_l$ and $\b_l$ are as follows:
\begin{align*}
\nabla_{\W_l}\mathscr{L} = &-\lambda_w\W_l
-\lambda_v\sum\limits_j\nabla_{\W_l}f_e(\X_{0,j*},\W^+)^T(f_e(\X_{0,j*},\W^+)^T-\v_j)\\
&-\lambda_n\sum\limits_j\nabla_{\W_l}f_r(\X_{0,j*},\W^+)(f_r(\X_{0,j*},\W^+)-\X_{c,j*})
\end{align*}
\begin{align*}
\nabla_{\b_l}\mathscr{L} = &-\lambda_w\b_l
-\lambda_v\sum\limits_j\nabla_{\b_l}f_e(\X_{0,j*},\W^+)^T(f_e(\X_{0,j*},\W^+)^T-\v_j)\\
&-\lambda_n\sum\limits_j\nabla_{\b_l}f_r(\X_{0,j*},\W^+)(f_r(\X_{0,j*},\W^+)-\X_{c,j*}).
\end{align*}
By alternating the update of $\U$, $\V$, $\W_l$, and $\b_l$, we can find a local optimum for $\mathscr{L}$. Several commonly used techniques such as using a momentum term may be applied to alleviate the local optimum problem.

\textbf{Prediction}: Let $D$ be the observed test data. Similar to \cite{CTR}, CDL uses the point estimates of $\u_i$, $\W^+$
%\footnote{** Typo fixed.}
and $\ep_j$ to calculate the predicted rating:
\begin{align}
E[\R_{ij}|D]\approx E[\u_i|D]^T(E[f_e(\X_{0,j*},\W^+)^T|D]+E[\ep_j|D]),\nonumber
\end{align}
where $E[\cdot]$ denotes the expectation operation.
In other words, we approximate the predicted rating as:%\footnote{*** Note that weird hyphenation problem: word-s.}
%\footnote{*** You haven't defined the terms in-matrix prediction and out-of-matrix prediction (even though they are used in the CTR paper). (** Modified. Since we performs in-matrix and out-of-matrix prediction together, there might not be necessary to treat them separately.)}
\begin{align}
\R^*_{ij}\approx(\u^*_j)^T(f_e(\X_{0,j*},{\W^+}^*)^T+\ep^*_j)=(\u^*_i)^T\v^*_j .\nonumber
\end{align}
Note that for any new item $j$ with no rating in the training data, its offset $\ep^*_j$ will be $\0$.

Recall that in CDL, the probabilistic SDAE and PMF work as the perception and task-specific components. As mentioned in Section~\ref{sec:bdl}, both components can take various forms, leading to different concrete models. For example, one can replace the probabilistic SDAE with a VAE or an NPN as the perception component~\cite{ColVAE}. It is also possible to use Bayesian PMF~\cite{BPMF} rather than PMF~\cite{PMF} as the task-specific component and thereby produce more robust predictions. 

In the following subsections, we provide several extensions of CDL from different perspectives. 

\subsubsection{Bayesian Collaborative Deep Learning}
Besides the MAP estimates, a sampling-based algorithm for the Bayesian treatment of CDL is also proposed in \cite{CDL}. This algorithm turns out to be a Bayesian and generalized version of BP. We list the key conditional densities as follows:

\textbf{For $\W^+$}:
We denote the concatenation of $\W_{l,*n}$ and $\b_l^{(n)}$ as $\W_{l,*n}^+$. Similarly, the concatenation of $\X_{l,j*}$ and $1$ is denoted as $\X_{l,j*}^+$. The subscripts of $\I$ are ignored. Then
\begin{align*}
p(\W_{l,*n}^+|\X_{l-1,j*},\X_{l,j*},\lambda_s)
\propto \  \NM(\W_{l,*n}^+|0,\lambda_w^{-1}\I) \cdot \NM(\X_{l,*n}|\sigma(\X_{l-1}^+\W_{l,*n}^+),\lambda_s^{-1}\I).
\end{align*}

\textbf{For $\X_{l,j*}$ ($l\neq L/2$)}:
Similarly, we denote the concatenation of $\W_{l}$ and $\b_l$ as $\W_{l}^+$ and have
\begin{align*}
&p(\X_{l,j*}|\W_l^+,\W_{l+1}^+,\X_{l-1,j*},\X_{l+1,j*}\lambda_s)\\
\propto \ & \NM(\X_{l,j*}|\sigma(\X_{l-1,j*}^+\W_l^+),\lambda_s^{-1}\I)\cdot
\NM(\X_{l+1,j*}|\sigma(\X_{l,j*}^+\W_{l+1}^+),\lambda_s^{-1}\I),
\end{align*}
where for the last layer ($l=L$) the second Gaussian would be $\NM(\X_{c,j*}|\X_{l,j*},\lambda_s^{-1}\I)$ instead.

\textbf{For $\X_{l,j*}$ ($l= L/2$)}:
Similarly, we have
\begin{align*}
&p(\X_{l,j*}|\W_l^+,\W_{l+1}^+,\X_{l-1,j*},\X_{l+1,j*},\lambda_s,\lambda_v,\v_j) \\
\propto \ &\NM(\X_{l,j*}|\sigma(\X_{l-1,j*}^+\W_l^+),\lambda_s^{-1}\I)\cdot
\NM(\X_{l+1,j*}|\sigma(\X_{l,j*}^+\W_{l+1}^+),\lambda_s^{-1}\I)\cdot \NM(\v_j|\X_{l,j*},\lambda_v^{-1}\I).
\end{align*}

\textbf{For $\v_j$}: The posterior 
% \begin{align*}
$
p(\v_j|\X_{L/2,j*},\R_{*j},\C_{*j},\lambda_v,\U) \propto\NM(\v_j|\X_{L/2,j*}^T,\lambda_v^{-1}\I)\prod\limits_i \NM(\R_{ij}|\u_i^T\v_j,\C_{ij}^{-1}).
$
% \end{align*}

\textbf{For $\u_i$}: The posterior 
% \begin{align*}
$
p(\u_i|\R_{i*},\V,\lambda_u,\C_{i*})\propto\NM(\u_i|0,\lambda_u^{-1}\I)\prod\limits_j\NM(\R_{ij}|\u_i^T\v_j,\C_{ij}^{-1}).
$
% \end{align*}

Interestingly, if $\lambda_s$ goes to infinity and adaptive rejection Metropolis sampling (which involves using the gradients of the objective function to approximate the proposal distribution) is used, the sampling for $\W^+$ turns out to be a \emph{Bayesian generalized} version of BP. Specifically, as Figure~\ref{fig:twonet}(right) shows, after getting the gradient of the loss function at one point (the red dashed line on the left), the next sample would be drawn in the region under that line, which is equivalent to a probabilistic version of BP. If a sample is above the curve of the loss function, a new tangent line (the black dashed line on the right) would be added to better approximate the distribution corresponding to the loss function. After that, samples would be drawn from the region under both lines. During the sampling, besides searching for local optima using the gradients (MAP), the algorithm also takes the variance into consideration. That is why it is called \emph{Bayesian generalized back-propagation} in \cite{CDL}.

% \begin{figure}[!tb]
% \begin{center}
% \includegraphics[height=4.0cm]{fig/sampling}
% \end{center}
% \vskip -0.3in
% \caption{Sampling as generalized BP.
% }
% \label{fig:sampling}
% \vskip -0.2in
% \end{figure}

\subsubsection{Marginalized Collaborative Deep Learning}
In SDAE, corrupted input goes through the encoder and decoder to recover the clean input. Usually, different epochs of training use different corrupted versions as input. Hence generally, SDAE needs to go through enough epochs of training to see sufficient corrupted versions of the input. Marginalized SDAE (mSDAE) \cite{mSDAE} seeks to avoid this by marginalizing out the corrupted input and obtaining closed-form solutions directly. In this sense, mSDAE is more computationally efficient than SDAE.

As mentioned in \cite{li2015deep}, using mSDAE instead of the Bayesian SDAE could lead to more efficient learning algorithms. For example, in \cite{li2015deep}, the objective when using a one-layer mSDAE can be written as follows:
\begin{align}
\mathscr{L}=
-\sum\limits_j\|\widetilde{\X}_{0,j*}\W_1-\overline{\X}_{c,j*}\|_2^2 -\sum\limits_{i,j}\frac{\C_{ij}}{2}(\R_{ij}-\u_i^T\v_j)^2 \nonumber
-\frac{\lambda_u}{2}\sum\limits_i \|\u_i\|_2^2
-\frac{\lambda_v}{2}\sum\limits_j\|\v_j^T\P_1-\X_{0,j*}\W_1\|_2^2,
\end{align}
where $\widetilde{\X}_{0,j*}$ is the collection of $k$ different corrupted versions of ${\X}_{0,j*}$ (a $k$-by-$B$ matrix) and $\overline{\X}_{c,j*}$ is the $k$-time repeated version of ${\X}_{c,j*}$ (also a $k$-by-$B$ matrix). $\P_1$ is the transformation matrix for item latent factors.

The solution for $\W_1$ would be $\W_1=E(\S_1)E(\Q_1)^{-1}$, where $\S_1=\overline{\X}_{c,j*}^T\widetilde{\X}_{0,j*}+\frac{\lambda_v}{2}\P_1^T\V\X_c$ and $\Q_1=\overline{\X}_{c,j*}^T\widetilde{\X}_{0,j*}+\frac{\lambda_v}{2}\X_c^T\X_c$. A solver for the expectation in the equation above is provided in \cite{mSDAE}. Note that this is a linear and one-layer case which can be generalized to the nonlinear and multi-layer case using the same techniques as in \cite{mSDAE,nmSDAE}.

Marginalized CDL's perception variables $\Om_p=\{\X_0,\X_c,\W_1\}$, its hinge variables $\Om_h=\{\V\}$, and its task variables $\Om_t=\{\P_1,\R,\U\}$.

\subsubsection{Collaborative Deep Ranking}
CDL assumes a collaborative filtering setting to model the ratings directly. Naturally, one can design a similar model to focus more on the ranking among items rather than exact ratings~\cite{yingcollaborative}.
% However, the output of recommender systems is often a ranked list, which means it would be more natural to use ranking rather than ratings as the objective. With this motivation, collaborative deep ranking (CDR) is proposed \cite{yingcollaborative} to jointly perform representation learning and collaborative ranking. 
The corresponding generative process is as follows:\\
\begin{compactenum}
\item For each layer $l$ of the SDAE network,
\begin{compactenum}
\item For each column $n$ of the weight matrix $\W_l$, draw
$\W_{l,*n} \sim \NM(\0,\lambda_w^{-1} \I_{K_l})$.
\item Draw the bias vector $\b_l \sim \NM(\0,\lambda_w^{-1} \I_{K_l})$.
\item For each row $j$ of $\X_l$, draw 
$
\X_{l,j*} \sim \NM(\sigma(\X_{l-1,j*}\W_l+\b_l),\lambda_s^{-1} \I_{K_l}).
$
\end{compactenum}
\item For each item $j$,
\begin{compactenum}
\item Draw a clean input $\X_{c,j*} \sim \NM(\X_{L,j*},\lambda_n^{-1} \I_{J}$).
\item Draw a latent item offset vector $\ep_j \sim \NM(\0,\lambda_v^{-1}\I_K)$ and then set the latent item vector to be:
$
\v_j=\ep_j+\X_{\frac{L}{2},j*}^T.
$
\end{compactenum}
\item For each user $i$,
\begin{compactenum}
\item Draw a latent user vector for each user $i$: 
$
\u_i \sim \NM(\0,\lambda_u^{-1}\I_K).
$
\item For each pair-wise preference $(j,k)\in\mathcal{P}_i$, where $\mathcal{P}_i=\{(j,k):\R_{ij}-\R_{ik}>0\}$, draw the preference:
$
\De_{ijk} \sim \NM(\u_i^T\v_j-\u_i^T\v_k,\C_{ijk}^{-1}).\\
$
\end{compactenum}
\end{compactenum}

Following the generative process above, the log-likelihood in Equation (\ref{eq:L}) becomes:
\begin{align}%\label{eq:L_ranking}
\mathscr{L}=&-\frac{\lambda_u}{2}\sum\limits_i \|\u_i\|_2^2
-\frac{\lambda_w}{2}\sum\limits_l(\|\W_l\|_F^2+\|\b_l\|_2^2)\nonumber-\frac{\lambda_v}{2}\sum\limits_j\|\v_j-f_e(\X_{0,j*},\W^+)^T\|_2^2 \nonumber \\
&-\frac{\lambda_n}{2}\sum\limits_j\|f_r(\X_{0,j*},\W^+)-\X_{c,j*}\|_2^2 \nonumber -\sum\limits_{i,j,k}\frac{\C_{ijk}}{2}(\De_{ijk}-(\u_i^T\v_j-\u_i^T\v_k))^2.
\end{align}
Similar algorithms can be used to learn the parameters in CDR. As reported in \cite{yingcollaborative}, using the ranking objective leads to significant improvement in the recommendation performance. Following the definition in Section \ref{sec:general}, CDR's perception variables $\Om_p=\{\{\W_l\},\{\b_l\},\{\X_l\},\X_c\}$, the hinge variables $\Om_h=\{\V\}$, and the task variables $\Om_t=\{\U,\De\}$.

\subsubsection{Collaborative Variational Autoencoders}

In CDL, the perception component takes the form of a probabilistic SDAE. Naturally, one can also replace the probabilistic SDAE in CDL with a VAE (introduced in Section~\ref{sec:vae}), as is done in collaborative variational autoencoders (CVAE)~\cite{ColVAE}. Specifically, CVAE with a inference network (encoder) denoted as $(f_{\mu}(\cdot), f_{s}(\cdot))$ and a generation network (decoder) denoted as $g(\cdot)$ assumes the following generative process:

\begin{compactenum}
% \item For each layer $l$ of the VAE encoder ($l=1,2,\dots,L$),
% \begin{compactenum}
% \item For each column $n$ of the weight matrix $\W_l$, draw
% $\W_{l,*n} \sim \NM(\0,\lambda_w^{-1} \I_{K_l})$.
% \item Draw the bias vector $\b_l \sim \NM(\0,\lambda_w^{-1} \I_{K_l})$.
% \item For each row $j$ of $\X_l$, draw 
% $
% \X_{l,j*} \sim \NM(\sigma(\X_{l-1,j*}\W_l+\b_l),\lambda_s^{-1} \I_{K_l}).
% $
% \end{compactenum}
\item For each item $j$,
\begin{compactenum}
\item Draw the latent item vector from the VAE inference network:
% \begin{align*}
$
\z_j \sim \NM(f_{\mu}(\X_{0, j*}), f_{s}(\X_{0, j*}))
$
% \end{align*}
\item Draw the latent item offset vector $\ep_j \sim \NM(\0,\lambda_v^{-1}\I_K)$ and then set the latent item vector:
$
\v_j=\ep_j+\z_j.
$
\item Draw the orignial input from the VAE generation network $\X_{0,j*} \sim \NM(g(\z_j),\lambda_n^{-1} \I_{B}$).
\end{compactenum}
\item Draw a latent user vector for each user $i$: 
$
\u_i \sim \NM(\0,\lambda_u^{-1}\I_K).
$
\item Draw a rating $\R_{ij}$ for each user-item pair $(i,j)$:
$
\R_{ij} \sim \NM(\u_i^T\v_j,\C_{ij}^{-1}).
$
\end{compactenum}
Similar to CDL, $\lambda_n$, $\lambda_u$, $\lambda_s$, and $\lambda_v$ are hyperparameters and $\C_{ij}$ is a confidence parameter ($\C_{ij} = a$ if $\R_{ij}=1$ and $\C_{ij}=b$ otherwise). Following \cite{ColVAE}, the ELBO similar to Equation~(\ref{eq:vae_elbo}) can be derived, using which one can train the model's parameters using BP and the reparameterization trick.  

The evolution from CDL to CVAE demonstrates the BDL framework's flexibility in terms of its components' specific forms. It is also worth noting that the perception component can be a recurrent version of probabilistic SDAE~\cite{CRAE} or VAE~\cite{VRNN,ColVAE} to handle raw sequential data, while the task-specific component can take more sophisticated forms to accommodate more complex recommendation scenarios (e.g., cross-domain recommendation). 

\subsubsection{Discussion}
Recommender systems are a typical use case for BDL in that they often require both thorough understanding of high-dimensional signals (e.g., text and images) and principled reasoning on the conditional dependencies among users/items/ratings.

In this regard, CDL, as an instantiation of BDL, is the first hierarchical Bayesian model to bridge the gap between state-of-the-art deep learning models and recommender systems. By performing deep learning collaboratively, CDL and its variants can simultaneously extract an effective deep feature representation from high-dimensional content and capture the similarity and implicit relationship between items (and users). The learned representation may also be used for tasks other than recommendation. Unlike previous deep learning models which use a simple target like classification \cite{DBLP:journals/acl/KalchbrennerGB14} and reconstruction \cite{DBLP:journals/jmlr/VincentLLBM10}, CDL-based models use CF as a more complex target in a probabilistic framework.

As mentioned in Section \ref{sec:intro}, \emph{information exchange} between two components is crucial for the performance of BDL. In the CDL-based models above, the exchange is achieved by assuming Gaussian distributions that connect the hinge variables and the variables in the perception component (drawing the hinge variable $\v_j \sim \NM(\X_{\frac{L}{2},j*}^T,\lambda_v^{-1}\I_K)$ in the generative process of CDL, where $\X_{\frac{L}{2}}$ is a perception variable), which is simple but effective and efficient in computation. Among the eight CDL-based models in Table \ref{table:summary}, six of them are HV models and the others are LV models, according to the definition of Section \ref{sec:general}. Since it has been verified that the HV CDL significantly outperforms its ZV counterpart \cite{CDL}, we can expect additional performance boost from the LV counterparts of the six HV models. 

Besides efficient \emph{information exchange}, the model designs also meet the independence requirement on the distribution concerning hinge variables discussed in Section \ref{sec:general} and are hence easily parallelizable. In some models to be introduced later, we will see alternative designs to enable efficient and independent information exchange between the two components of BDL. 

Note that BDL-based models above use typical static Bayesian networks as their task-specific components. Although these are often sufficient for most use cases, it is possible for the task-specific components to take the form of deep Bayesian networks such as BIN~\cite{BIN}. This allows the models to handle highly nonlinear interactions between users and items if necessary. One can also use stochastic processes (or dynamic Bayesian networks in general) to explicitly model users purchase or clicking behaviors. For example, it is natural to model a user's purchase of groceries as a Poisson process. In terms of perception components, one can also replace the pSDAE, mSDAE, or VAE above with their convolutional or recurrent counterparts (see Section~\ref{sec:cnn} and Section~\ref{sec:rnn}), as is done in Collaborative Knowledge Base Embedding (CKE)~\cite{CKE} or Collaborative Recurrent Autoencoders (CRAE)~\cite{CRAE}, respectively. Note that for the convolutional or recurrent perception components to be compatible with the task-specific component (which is inherently probabilistic), ideally one would need to formulate probabilistic versions of CNN or RNN as well. Readers are referred to \cite{CKE} and \cite{CRAE} for more details.

% \textred{BIN as deep Bayesian MF, stochastic process to model users' purchase or clicking behaviors e.g., as a Poisson process}

In summary, this subsection discusses BDL's applications on supervised leraning, using recommender systems as an example. Section~\ref{sec:topic_models} below will cover BDL's applications on unsupervised learning. 

\subsection{Unsupervised Bayesian Deep Learning for Topic Models}\label{sec:topic_models}
To demonstrate how BDL can also be applied to unsupervised learning, we review some examples of BDL-based topic models in this section. These models combine the merits of PGM (which naturally incorporates the probabilistic relations among variables) and NN (which learns deep representations efficiently), leading to significant performance boost. In the case of unsupervised learning, the `task' for a task-specific component is to describe/characterize the conditional dependencies in the BDL model, thereby improving its interpretability and genearalizability. This is different from the supervised learning setting where the `task' is simply to `match the target'.

\subsubsection{Relational Stacked Denoising Autoencoders as Topic Models}\label{sec:rsdae}
As a BDL-based topic model, relational stacked denoising autoencoders (RSDAE) essentially tries to learn a hierarchy of topics (or latent factors) while enforcing relational (graph) constraints under an unsupervised learning setting. 

\textbf{Problem Statement and Notation}:
Assume we have a set of items~(articles or movies) $\X_c$, with $\X_{c,j*}^T \in \RB^B$ denoting the content (attributes) of item $j$. Besides, we use $\I_K$ to denote a $K$-dimensional identity matrix and $\S=[\s_1,\s_2,\cdots,\s_J]$ to denote the \emph{relational latent matrix} with $\s_j$ representing the relational properties of item $j$.
%\footnote{** Add notation for $\S$}

From the perspective of SDAE, the $J$-by-$B$
%\footnote{***Has the symbol $S$ been defined? ** Use B instead.}
matrix $\X_c$ represents the clean input to the SDAE and the noise-corrupted matrix of the same size is denoted by $\X_0$. Besides, we denote the output of layer $l$ of the SDAE, a $J$-by-$K_l$ matrix, by $\X_l$. Row $j$ of $\X_l$ is denoted by $\X_{l,j*}$, $\W_l$ and $\b_l$ are the weight matrix and bias vector of layer $l$, $\W_{l,*n}$ denotes column $n$ of $\W_l$, and $L$ is the number of layers. As a shorthand, we refer to the collection of weight matrices and biases in all layers as $\W^+$. Note that an $L/2$-layer SDAE corresponds to an $L$-layer network.
% can be merged with the CDL part

\begin{figure} [tbp]
\begin{center}
\vskip -0.2in
  \begin{tabular}{ccc}
   \includegraphics*[height=35mm]{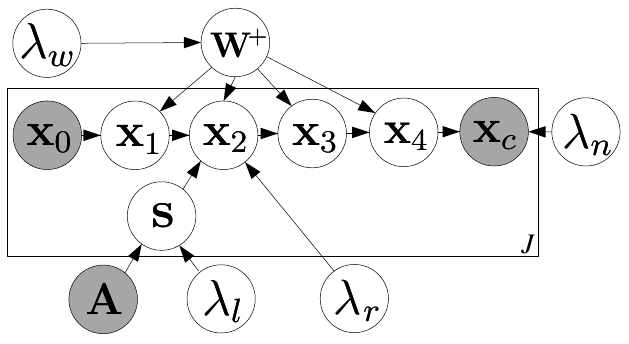}
  \end{tabular} \vskip -0.3cm
  \vskip -0.1in
\caption{Graphical model of RSDAE for $L=4$. $\lambda_s$ is omitted here to prevent clutter.} \label{fig:rsdae}
\end{center}
\vskip -0.3in
\end{figure}

\textbf{Model Formulation}:
In RSDAE, the perception component takes the form of a probabilistic SDAE (introduced in Section~\ref{sec:psdae}) as a building block. At a higher level, RSDAE is formulated as a novel probabilistic model which seamlessly integrates a hierarchy of latent factors and the relational information available. This way, the model can simultaneously learn the feature representation from the content information and the relation among items~\cite{RSDAE}. The graphical model for RSDAE is shown in Figure~\ref{fig:rsdae}, and the generative process is listed as follows:\\

\begin{compactenum}
\item Draw the relational latent matrix $\S$  from a \emph{matrix-variate normal distribution}~\cite{gupta2000matrix}: 
\begin{align}
\S\sim\NM_{K,J}(0,\I_K\otimes (\lambda_l \mathscr{L}_a)^{-1}).\label{eq:nm}
\end{align}
\item For layer $l$ of the SDAE where $l=1,2,\dots,\frac{L}{2}-1$,
\begin{compactenum}
\item For each column $n$ of the weight matrix $\W_l$, draw $\W_{l,*n} \sim \NM(0,\lambda_w^{-1} \I_{K_l})$.
\item Draw the bias vector $\b_l \sim \NM(0,\lambda_w^{-1} \I_{K_l})$.
\item For each row $j$ of $\X_l$, draw 
$
\X_{l,j*} \sim \NM(\sigma(\X_{l-1,j*}\W_l+\b_l),\lambda_s^{-1} \I_{K_l}).
$
\end{compactenum}
\item For layer $\frac{L}{2}$ of the SDAE, draw the representation vector for item $j$ from the product of two Gaussians~(PoG)~\cite{DBLP:journals/csl/GalesA06}:
\begin{align}
\X_{\frac{L}{2},j*}\sim\mbox{PoG}(\sigma(\X_{\frac{L}{2}-1,j*}\W_l+\b_l),\s_j^T,\lambda_s^{-1} \I_K,\lambda_r^{-1}\I_K). \label{eq:pog}
\end{align}
\item For layer $l$ of the SDAE where $l=\frac{L}{2}+1,\frac{L}{2}+2,\dots,L$,
\begin{compactenum}
\item For each column $n$ of the weight matrix $\W_l$, draw $\W_{l,*n} \sim \NM(0,\lambda_w^{-1} \I_{K_l})$.
\item Draw the bias vector $\b_l \sim \NM(0,\lambda_w^{-1} \I_{K_l})$.
\item For each row $j$ of $\X_l$, draw 
$\X_{l,j*} \sim \NM(\sigma(\X_{l-1,j*}\W_l+\b_l),\lambda_s^{-1} \I_{K_l}).$
\end{compactenum}
\item For each item $j$, draw a clean input $\X_{c,j*} \sim \NM(\X_{L,j*},\lambda_n^{-1}\I_{B}).$\\
\end{compactenum}
Here $K=K_{\frac{L}{2}}$ is the dimensionality of the learned representation vector for each item, $\S$ denotes the $K\times J$ relational latent matrix in which column $j$ is the \emph{relational latent vector} $\s_j$ for item $j$. Note that $\NM_{K,J}(0,\I_K\otimes (\lambda_l \mathscr{L}_a)^{-1})$ in~Equation~(\ref{eq:nm})
%\footnote{*** Incorrect equation cited. ** Problem fixed.}
is a matrix-variate normal distribution defined as in~\cite{gupta2000matrix}:
%\footnote{*** $p(\S)$ ** Modified.}
\begin{align}\label{eq:mvd}
p(\S) = \NM_{K,J}(0,\I_K\otimes (\lambda_l \mathscr{L}_a)^{-1}) 
=\frac{\exp\{\tr[-\frac{\lambda_l }{2}\S \mathscr{L}_a \S^T]\}}{(2\pi)^{JK/2}|\I_K|^{J/2}|\lambda_l \mathscr{L}_a|^{-K/2}},
\end{align}
where the operator $\otimes$ denotes the Kronecker product of two matrices \cite{gupta2000matrix}, $\tr(\cdot)$ denotes the trace of a matrix, and $\mathscr{L}_a$ is the Laplacian matrix incorporating the relational information.
%\footnote{*** Social network? ** changed to relational}
$\mathscr{L}_a=\D-\A$, where $\D$ is a diagonal matrix whose diagonal elements $\D_{ii}=\sum_j\A_{ij}$ and $\A$ is the adjacency matrix representing the relational information
%\footnote{** Changed.}
with binary entries indicating the links~(or relations) between items. $\A_{jj'}=1$ indicates that there is a link between item $j$ and item $j'$ and $\A_{jj'}=0$ otherwise. $\mbox{PoG}(\sigma(\X_{\frac{L}{2}-1,j*}\W_l+\b_l),\s_j^T,\lambda_s^{-1} \I_K,\lambda_r^{-1}\I_K)$ denotes the product of the Gaussian $\NM(\sigma(\X_{\frac{L}{2}-1,j*}\W_l+\b_l),\lambda_s^{-1} \I_K)$ and the Gaussian $\NM(\s_j^T,\lambda_r^{-1}\I_K)$, which is also a Gaussian \cite{DBLP:journals/csl/GalesA06}.
%The resulting Gaussian is $\NM(\mu_{sr},\lambda_{sr}^{-1}I_K)$ with
%\begin{align}
%\mu_{sr}&=\frac{\sigma(\X_{\frac{L}{2}-1,j*}\W_l+\b_l)\lambda_s+\s_j^T\lambda_r}{\lambda_v+\lambda_r}, \nonumber \\
%\lambda_{sr} &=\frac{\lambda_s\lambda_r}{\lambda_s+\lambda_r}. \nonumber
%\end{align}

According to the generative process above, maximizing the posterior probability is equivalent to maximizing the joint log-likelihood of $\{\X_l\}$, $\X_c$, $\S$, $\{\W_l\}$, and $\{\b_l\}$ given $\lambda_s$, $\lambda_w$, $\lambda_l$, $\lambda_r$, and $\lambda_n$:
\begin{align}
\mathscr{L}=&-\frac{\lambda_l}{2}\tr(\S\mathscr{L}_a \S^T)-\frac{\lambda_r}{2}\sum\limits_j\|(\s_j^T-\X_{\frac{L}{2},j*})\|_2^2
-\frac{\lambda_w}{2}\sum\limits_l(\|\W_l\|_F^2+\|\b_l\|_2^2) \nonumber \\
&-\frac{\lambda_n}{2}\sum\limits_j\|\X_{L,j*}-\X_{c,j*}\|_2^2 
-\frac{\lambda_s}{2}\sum\limits_l\sum\limits_j\|\sigma(\X_{l-1,j*}\W_l+\b_l)-\X_{l,j*}\|_2^2.\nonumber
\end{align}

Similar to the pSDAE, taking $\lambda_s$ to infinity, the joint log-likelihood becomes:
\begin{align}\label{eq:rsdae}
\mathscr{L}=-\frac{\lambda_l}{2}\tr(\S\mathscr{L}_a \S^T)-\frac{\lambda_r}{2}\sum\limits_j\|(\s_j^T-\X_{\frac{L}{2},j*})\|_2^2
-\frac{\lambda_w}{2}\sum\limits_l(\|\W_l\|_F^2+\|\b_l\|_2^2) 
-\frac{\lambda_n}{2}\sum\limits_j\|\X_{L,j*}-\X_{c,j*}\|_2^2,
\end{align}
where $\X_{l,j*}=\sigma(\X_{l-1,j*}\W_l+\b_l)$. Note that the first term $-\frac{\lambda_l}{2}\tr(\S\mathscr{L}_a \S^T)$ corresponds to $\log p(\S)$
%\footnote{*** $\log p(\S)$ ** Modified.}
in the matrix-variate distribution in Equation (\ref{eq:mvd}).
%\footnote{*** ?? ** Typo fixed.}
Besides, by simple manipulation, we have
%\begin{align}
%\tr(\S \mathscr{L}_a \S^T) &=\frac{1}{2}\sum\limits_{j=1}^J\sum\limits_{j'=1}^J \A_{jj'}||\S_{*j}-\S_{*j'}||^2 \\ \nonumber
%&=\frac{1}{2}\sum\limits_{j=1}^J\sum\limits_{j'=1}^J[\A_{jj'}\sum\limits_{k=1}^K(\S_{kj}-\S_{kj'})^2]\\ \nonumber
%&=\frac{1}{2}\sum\limits_{k=1}^{K}[\sum\limits_{j=1}^J\sum\limits_{j'=1}^J \A_{jj'}(\S_{kj}-\S_{kj'})^2]\\ \nonumber
%&=\sum\limits_{k=1}^K \S_{k*}^T\mathscr{L}_a\S_{k*}, \nonumber
%\end{align}
$
\tr(\S \mathscr{L}_a \S^T)
=\sum\limits_{k=1}^K \S_{k*}^T\mathscr{L}_a\S_{k*}
$,
where $\S_{k*}$ denotes the $k$-th row of $\S$. As we can see, maximizing $-\frac{\lambda_l}{2}\tr(\S^T\mathscr{L}_a\S)$ is equivalent to making $\s_j$ closer to $\s_{j'}$ if item $j$ and item $j'$ are linked (namely $\A_{jj'}=1$)~\cite{CTRSR}.

In RSDAE, the perception variables $\Om_p=\{\{\X_l\},\X_c,\{\W_l\},\{\b_l\}\}$, the hinge variables $\Om_h=\{\S\}$, and the task variables $\Om_t=\{\A\}$.

\textbf{Learning and Inference}:
\cite{RSDAE} provides an EM-style algorithm for MAP estimation. Below we review some of the key steps.

For the E step, the challenge lies in the inference of the relational latent matrix $\S$.  We first fix all rows of $\S$ except the $k$-th one $\S_{k*}$ and then update $\S_{k*}$. Specifically, we take the gradient of $\mathscr{L}$ with respect to $\S_{k*}$, set it to 0, and get the following linear system:
\begin{align}
(\lambda_l\mathscr{L}_a+\lambda_r\I_J)\S_{k*}=\lambda_r\X_{\frac{L}{2},*k}^T.
\end{align}
A naive approach is to solve the linear system by setting $\S_{k*}=\lambda_r(\lambda_l\mathscr{L}_a+\lambda_r\I_J)^{-1}\X_{\frac{L}{2},*k}^T$. Unfortunately, the complexity is $O(J^3)$ for one single update. Similar to \cite{DBLP:conf/ijcai/LiY09}, the steepest descent method \cite{techreport/Shewchuk94} is used to iteratively update $\S_{k*}$:
\begin{align}
\S_{k*}(t+1)\leftarrow \S_{k*}(t)+\delta(t)r(t), \;\;\;
r(t)\leftarrow \lambda_r\X_{\frac{L}{2},*k}^T-(\lambda_l\mathscr{L}_a+\lambda_r\I_J)\S_{k*}(t), \;\;\;
\delta(t)\leftarrow \frac{r(t)^Tr(t)}{r(t)^T(\lambda_l\mathscr{L}_a+\lambda_r\I_J)r(t)}.\nonumber
\end{align}
% \begin{align}
% \S_{k*}(t+1)&\leftarrow \S_{k*}(t)+\delta(t)r(t)\nonumber\\
% r(t)&\leftarrow \lambda_r\X_{\frac{L}{2},*k}^T-(\lambda_l\mathscr{L}_a+\lambda_r\I_J)\S_{k*}(t)\nonumber\\
% \delta(t)&\leftarrow \frac{r(t)^Tr(t)}{r(t)^T(\lambda_l\mathscr{L}_a+\lambda_r\I_J)r(t)}.\nonumber
% \end{align}
As discussed in \cite{DBLP:conf/ijcai/LiY09}, the steepest descent method dramatically reduces the computation cost in each iteration from $O(J^3)$ to $O(J)$.

The M step involves learning $\W_l$ and $\b_l$ for each layer using the back-propagation algorithm given $\S$. By alternating the update of $\S$, $\W_l$, and $\b_l$, a local optimum for $\mathscr{L}$ can be found. Also, techniques such as including a momentum term may help to avoid being trapped in a local optimum.

\subsubsection{Deep Poisson Factor Analysis with Sigmoid Belief Networks}
The Poisson distribution with support over nonnegative integers is known as a natural choice to model counts. It is, therefore, desirable to use it as a building block for topic models, which are generally interested in word counts~\cite{LDA}. With this motivation, \cite{PFA} proposed a model, dubbed Poisson factor analysis (PFA), for latent nonnegative matrix factorization via Poisson distributions. 

\textbf{Poisson Factor Analysis}:
PFA assumes a discrete $P$-by-$N$ matrix $\X$ containing word counts of $N$ documents with a vocabulary size of $P$ \cite{PFA,DPFA}. In a nutshell, PFA can be described using the equation $\X\sim \mbox{Pois}(\Ph(\Tha\circ\H))$, where $\Ph$ (of size $P$-by-$K$ where $K$ is the number of topics) denotes the factor loading matrix in factor analysis with the $k$-th column $\ph_k$ encoding the importance of each word in topic $k$. The $K$-by-$N$ matrix $\Tha$ is the factor score matrix with the $n$-th column $\tha_n$ containing topic proportions for document $n$. The $K$-by-$N$ matrix $\H$ is a latent binary matrix with the $n$-th column $\h_n$ defining a set of topics associated with document $n$. %Note that the superscript $(1)$ indexes a layer, which is irrelevant for one-layer PFA.

Different priors correspond to different models. For example, Dirichlet priors on $\ph_k$ and $\tha_n$ with an all-one matrix $\H$ would recover LDA \cite{LDA} while a beta-Bernoulli prior on $\h_n$ leads to the NB-FTM model in \cite{DBLP:journals/pami/ZhouC15}. In \cite{DPFA}, a deep-structured prior based on sigmoid belief networks (SBN) \cite{neal1992} (an MLP variant with binary hidden units) is imposed on $\h_n$ to form a deep PFA model for topic modeling.

\textbf{Deep Poisson Factor Analysis}:
In the deep PFA model \cite{DPFA}, the generative process can be summarized as follows:
% \begin{align}
% \ph_k&\sim\mbox{Dir}(a_{\phi},\dots,a_{\phi}), \;\;\;\;
% \theta_{kn}\sim\mbox{Gamma}(r_k,\frac{p_n}{1-p_n}), \;\;\;\; 
% r_k\sim\mbox{Gamma}(\gamma_0,\frac{1}{c_0}), \;\;\;\; 
% \gamma_0\sim\mbox{Gamma}(e_0,\frac{1}{f_0}),\nonumber\\
% h_{k_Ln}^{(L)}&\sim\mbox{Ber}(\sigma(b_{k_L}^{(L)})),\label{eq:sbn_top}\\
% h_{k_ln}^{(l)}&\sim\mbox{Ber}(\sigma({\w_{k_l}^{(l)}}^T\h_n^{(l+1)}+b_{k_l}^{(l)})),\label{eq:sbn_others}\\
% x_{pnk}&\sim\mbox{Pois}(\phi_{pk}\theta_{kn}h_{kn}^{(1)}), \label{eq:sbn_x} \\
% x_{pn}&=\sum\limits_{k=1}^Kx_{pnk},\nonumber
% \end{align}
\begin{align}
\ph_k&\sim\mbox{Dir}(a_{\phi},\dots,a_{\phi}), \;\;\;\;
\theta_{kn}\sim\mbox{Gamma}(r_k,\frac{p_n}{1-p_n}), \;\;\;\; 
r_k\sim\mbox{Gamma}(\gamma_0,\frac{1}{c_0}), \;\;\;\; 
\gamma_0\sim\mbox{Gamma}(e_0,\frac{1}{f_0}),\nonumber\\
h_{k_Ln}^{(L)}&\sim\mbox{Ber}(\sigma(b_{k_L}^{(L)})), \;\;\;\; 
h_{k_ln}^{(l)}\sim\mbox{Ber}(\sigma({\w_{k_l}^{(l)}}^T\h_n^{(l+1)}+b_{k_l}^{(l)})), \;\;\;\; 
x_{pnk}\sim\mbox{Pois}(\phi_{pk}\theta_{kn}h_{kn}^{(1)}),  \;\;\;\; 
x_{pn}=\sum\limits_{k=1}^Kx_{pnk}, \label{eq:sbn_all}
\end{align}
where $L$ is the number of layers in SBN, which corresponds to Equation (\ref{eq:sbn_all}). $x_{pnk}$ is the count of word $p$ that comes from topic $k$ in document $n$.

In this model, the perception variables $\Om_p=\{\{\H^{(l)}\},\{\W_l\},\{\b_l\}\}$, the hinge variables $\Om_h=\{\X\}$, and the task variables $\Om_t=\{\{\ph_k\},\{r_k\},\Tha,\gamma_0\}$. $\W_l$ is the weight matrix containing columns of $\w_{k_l}^{(l)}$ and $\b_l$ is the bias vector containing entries of $b_{k_l}^{(l)}$ in Equation (\ref{eq:sbn_all}).

\textbf{Learning Using Bayesian Conditional Density Filtering}:
Efficient learning algorithms are needed for Bayesian treatments of deep PFA. \cite{DPFA} proposed to use an online version of MCMC called Bayesian conditional density filtering (BCDF) to learn both the global parameters $\Ps_g=(\{\ph_k\},\{r_k\},\gamma_0,\{\W_l\},\{\b_l\})$ and the local variables $\Ps_l=(\Tha,\{\H^{(l)}\})$. The key conditional densities used for the Gibbs updates are as follows:
\begin{align*}
x_{pnk}|-&\sim\mbox{Multi}(x_{pn};\zeta_{pn1},\dots,\zeta_{pnK}), \;\;\;\;
\ph_k|-\sim\mbox{Dir}(a_{\phi}+x_{1\cdot k},\dots,a_{\phi}+x_{P\cdot k}), \\
\theta_{kn}|-&\sim\mbox{Gamma}(r_kh_{kn}^{(1)}+x_{\cdot nk},p_n), \;\;\;\;
h_{kn}^{(1)}|-\sim \delta(x_{\cdot nk}=0)\mbox{Ber}(\frac{\widetilde{\pi}_{kn}}{\widetilde{\pi}_{kn}+(1-\pi_{kn})})+\delta(x_{\cdot nk}>0),
\end{align*}
where $\widetilde{\pi}_{kn}=\pi_{kn}(1-p_n)^{r_k}$, $\pi_{kn}=\sigma((\w_k^{(1)})^T\h_n^{(2)}+c_k^{(1)})$, $x_{\cdot nk}=\sum\limits_{p=1}^P x_{pnk}$, $x_{p \cdot k}=\sum\limits_{n=1}^N x_{pnk}$, and $\zeta_{pnk}\propto\phi_{pk}\theta_{kn}$. For the learning of $h_{kn}^{(l)}$ where $l>1$, the same techniques as in \cite{DBLP:conf/aistats/GanHCC15} can be used.

\textbf{Learning Using Stochastic Gradient Thermostats}:
An alternative way of learning deep PFA is through \emph{stochastic gradient N\'ose-Hoover thermostats} (SGNHT), which is more accurate and scalable. SGNHT is a generalization of the \emph{stochastic gradient Langevin dynamics} (SGLD) \cite{SGLD} and the \emph{stochastic gradient Hamiltonian Monte Carlo} (SGHMC)~\cite{SGHMC}. Compared with the previous two, SGNHT introduces momentum variables into the system, helping the system to jump out of local optima. Specifically, the following stochastic differential equations (SDE) can be used:
\begin{align*}
d\Ps_g=\v dt, \;\;\;\;
d\v=\widetilde{f}(\Ps_g)dt-\xi\v dt+\sqrt{D}d\mathcal{W}, \;\;\;\; 
d\xi=(\frac{1}{M}\v^T\v-1)dt,
\end{align*}
where $\widetilde{f}(\Ps_g)=-\nabla_{\Ps_g}\widetilde{U}(\Ps_g)$ and $\widetilde{U}(\Ps_g)$ is the negative log-posterior of the model. $t$ indexes time and $\mathcal{W}$ denotes the standard Wiener process. $\xi$ is the thermostats variable to make sure the system has a constant temperature. $D$ is the injected variance which is a constant. To speed up convergence, the SDE is generalized to:
\begin{align*}
d\Ps_g=\v dt, \;\;\;\; 
d\v=\widetilde{f}(\Ps_g)dt-\Xii\v dt+\sqrt{D}d\mathcal{W}, \;\;\;\; 
d\Xii=(\q-\I)dt,
\end{align*}
where $\I$ is the identity matrix, $\Xii=\mbox{diag}(\xi_1,\dots,\xi_M)$, $\q=\mbox{diag}(v_1^2,\dots,v_M^2)$, and $M$ is the dimensionality of the parameters.

SGNHT, SGLD, and SGHMC all belong to a larger class of sampling algorithms called hybrid Monte Carlo (HMC)~\cite{PRML}. The idea is to leverage an analogy with physical systems to guide transitions of system states. Compared to the Metropolis algorithm, HMC can make much larger changes to system states while keeping a small rejection probability. For more details, we refer readers to \cite{PRML,neal2011mcmc}. 

\subsubsection{Deep Poisson Factor Analysis with Restricted Boltzmann Machine}
The deep PFA model above uses SBN as a perception component. Similarly, one can replace SBN with RBM~\cite{RBM} (discussed in Section~\ref{sec:rbm}) to achieve comparable performance. With RBM as the perception component, Equation (\ref{eq:sbn_all}) becomes conditional distributions similar to Equation~(\ref{eq:rbm_conditional}) with the following energy \cite{RBM}:
\begin{align*}
E(\h_n^{(l)},\h_n^{(l+1)})=-(\h_n^{(l)})^T\c^{(l)}-(\h_n^{(l)})^T\W^{(l)}\h_n^{(l+1)}
-(\h_n^{(l+1)})^T\c^{(l+1)}.
\end{align*}

Similar learning algorithms as the deep PFA with SBN can be used. Specifically, the sampling process would alternate between $\{\{\ph_k\},\{\gamma_k\},\gamma_0\}$ and $\{\{\W^{(l)}\},\{\c^{(l)}\}\}$. The former involves similar conditional density as the SBN-based DPFA. The latter is RBM's parameters and can be updated using the \emph{contrastive divergence} algorithm.

\subsubsection{Discussion}
Here we choose topic models as an example application to demonstrate how BDL can be applied in the unsupervised learning setting. In BDL-based topic models, the perception component is responsible for inferring the topic hierarchy from documents, while the task-specific component is in charge of modeling the word generation, topic generation, word-topic relation, or inter-document relation. The synergy between these two components comes from the bidirectional interaction between them. On one hand, knowledge on the topic hierarchy facilitates accurate modeling of words and topics, providing valuable information for learning inter-document relation. On the other hand, accurately modeling the words, topics, and inter-document relation can help discover the topic hierarchy and learn compact latent factors for documents.

It is worth noting that the \emph{information exchange} mechanism in some BDL-based topic models is different from that in Section \ref{sec:recsys}. For example, in the SBN-based DPFA model, the exchange is natural since the bottom layer of SBN, $\H^{(1)}$, and the relationship between $\H^{(1)}$ and $\Om_h=\{\X\}$ are both inherently probabilistic, as shown in Equation (\ref{eq:sbn_all}), which means additional assumptions on the distribution are not necessary. The SBN-based DPFA model is equivalent to assuming that $\H$ in PFA is generated from a Dirac delta distribution (a Gaussian distribution with zero variance) centered at the bottom layer of the SBN, $\H^{(1)}$. Hence both DPFA models in Table \ref{table:summary} are ZV models, according to the definition in Section \ref{sec:general}. It is worth noting that RSDAE is an HV model (see Equation (\ref{eq:pog}), where $\S$ is the hinge variable and the others are perception variables), and naively modifying this model to be its ZV counterpart would violate the i.i.d. requirement in Section~\ref{sec:general}.

Similar to Section~\ref{sec:recsys}, BDL-based topic models above use typical static Bayesian networks as task-specific components. Naturally, one can choose to use other forms of task-specific components. For example, it is straightforward to replace the relational prior of RSDAE in Section~\ref{sec:rsdae} with a stochastic process (e.g., a Wiener process as in \cite{cDTM}) to model the evolution of the topic hierarchy over time.

\subsection{Bayesian Deep Representation Learning for Control}\label{sec:control}
In Section~\ref{sec:recsys} and Section~\ref{sec:topic_models}, we covered how BDL can be applied in the supervised and unsupervised learning settings, respectively. In this section, we will discuss how BDL can help representation learning in general, using \emph{control} as an example application.

As mentioned in Section \ref{sec:intro}, Bayesian deep learning can also be applied to the control of nonlinear dynamical systems from raw images. Consider controlling a complex dynamical system according to the live video stream received from a camera. One way of solving this control problem is by iteration between two tasks, perception from raw images and control based on dynamic models. The perception task can be taken care of using multiple layers of simple nonlinear transformation (deep learning) while the control task usually needs more sophisticated models like hidden Markov models and Kalman filters \cite{harrison1999bayesian,DBLP:conf/uai/MatsubaraGK14}. To enable an effective iterative process between the perception task and the control task, we need two-way information exchange between them. The perception component would be the basis on which the control component estimates its states and on the other hand, the control component with a dynamic model built in would be able to predict the future trajectory (images) by reversing the perception process \cite{watter2015embed}.

As one of the pioneering works in this direction, \cite{watter2015embed} posed this task as a representation learning problem and proposed a model called \emph{Embed to Control} to take into account the feedback loop mentioned above during representation learning. Essentially, the goal is to learn representations that (1) capture semantic information from raw images/videos and (2) preserve local linearity in the state space for convenient control. This is not possible without the BDL framework since the perception component guarantees the first sub-goal while the task-specific component guarantees the second. Below we start with some preliminaries on stochastic optimal control and then introduce the BDL-based model for representation learning. 

%This problem can be transformed into iteratively perform two tasks

\subsubsection{Stochastic Optimal Control}
Following \cite{watter2015embed}, we consider the stochastic optimal control of an unknown dynamical system as follows:
\begin{align}
\z_{t+1}=f(\z_t,\u_t)+\xii,\ \xii\sim\NM(0,\Si_{\xi}),\label{eq:L_control}
\end{align}
where $t$ indexes the time steps and $\z_t\in\mathbb{R}^{n_z}$ is the latent states. $\u_t\in\mathbb{R}^{n_u}$ is the applied control at time $t$ and $\xii$ denotes the system noise. Equivalently, the equation above can be written as $P(\z_{t+1}|\z_t,\u_t)=\NM(\z_{t+1}|f(\z_t,\u_t),\Si_{\xi})$. Hence we need a mapping function to map the corresponding raw image $\x_t$ (observed input) into the latent space:
\begin{align*}
\z_t=m(\x_t)+\Ome,\ \Ome\sim\NM(0,\Si_{\omega}),
\end{align*}
where $\Ome$ is the corresponding system noise. Similarly the equation above can be rewritten as $\z_t\sim\NM(m(\x_t),\Si_{\omega})$. If the function $f$ is given, finding optimal control for a trajectory of length $T$ in a dynamical system amounts to minimizing the following cost:
\begin{align}
J(\z_{1:T},\u_{1:T})=\mathbb{E}_{\z}(c_T(\z_T,\u_T)+\sum\limits_{t_0}^{T-1} c(\z_t,\u_t)) \label{eq:control_cost},
\end{align}
where $c_T(\z_T,\u_T)$ is the terminal cost and $c(\z_t,\u_t)$ is the instantaneous cost. $\z_{1:T}=\{\z_1,\dots,\z_T\}$ and $\u_{1:T}=\{\u_1,\dots,\u_T\}$ are the state and action sequences, respectively. For simplicity we can let $c_T(\z_T,\u_T)=c(\z_T,\u_T)$ and use the following quadratic cost 
% \begin{align*}
$
c(\z_t,\u_t)=(\z_t-\z_{goal})^T\R_z(\z_t-\z_{goal})+\u_t^T\R_u\u_t,
$
% \end{align*}
where $\R_z\in\mathbb{R}^{n_z\times n_z}$ and $\R_u\in\mathbb{R}^{n_u\times n_u}$ are the weighting matrices. $\z_{goal}$ is the target latent state that should be inferred from the raw images (observed input). Given the function $f$, $\overline{\z}_{1:T}$ (current estimates of the optimal trajectory), and $\overline{\u}_{1:T}$ (the corresponding controls), the dynamical system can be linearized as:
\begin{align}
\z_{t+1}=\A(\overline{\z}_t)\z_t+\B(\overline{\z}_t)\u_{t}+\oo(\overline{\z}_t)+\Ome,\ \Ome\sim\NM(0,\Si_{\omega}),\label{eq:transition}
\end{align}
where $\A(\overline{\z}_t)=\frac{\partial f(\overline{\z}_t,\overline{\u}_t)}{\partial \overline{\z}_t}$ and $\B(\overline{\z}_t)=\frac{\partial f(\overline{\z}_t,\overline{\u}_t)}{\partial \overline{\u}_t}$ are local Jacobians. $\oo(\overline{\z}_t)$ is the offset.

\subsubsection{BDL-Based Representation Learning for Control}
To minimize the function in Equation (\ref{eq:control_cost}), we need three key components: an encoding model to encode $\x_t$ into $\z_t$, a transition model to infer $\z_{t+1}$ given $(\z_t,\u_t)$, and a reconstruction model to reconstruct $\x_{t+1}$ from the inferred $\z_{t+1}$.

\textbf{Encoding Model}:
An encoding model $Q_{\phi}(Z|X)=\NM(\muu_t,\mbox{diag}(\si_t^2))$, where the mean $\muu_t\in\mathbb{R}^{n_z}$ and the diagonal covariance $\Si_t=\mbox{diag}(\si_t^2)\in\mathbb{R}^{n_z\times n_z}$, encodes the raw images $\x_t$ into latent states $\z_t$. Here,
\begin{align}
\muu_t&=\W_{\mu}h_{\phi}^{\text{enc}}(\x_t)+\b_{\mu}, \;\;\;\;
\log \si_t=\W_{\sigma}h_{\phi}^{\text{enc}}(\x_t)+\b_{\sigma}, \label{eq:encode_all}
\end{align}
where $h_{\phi}(\x_t)^{\text{enc}}$ is the output of the encoding network with $\x_t$ as its input.

\textbf{Transition Model}:
A transition model like Equation (\ref{eq:transition}) infers $\z_{t+1}$ from $(\z_t,\u_t)$. If we use $\widetilde{Q}_{\psi}(\widetilde{Z}|Z,\u)$ to denote the approximate posterior distribution to generate $\z_{t+1}$, the generative process of the full model would be:
% \begin{align}
% \z_t&\sim Q_{\phi}(Z|X)=\NM(\muu_t,\Si_t), \label{eq:encode}\\
% \widetilde{\z}_{t+1}&\sim \widetilde{Q}_{\psi}(\widetilde{Z}|Z,\u)=\NM(\A_t\muu_t+\B_t\u_t+\oo_t,\C_t), \label{eq:trans}\\
% \widetilde{\x}_t,\widetilde{\x}_{t+1}&\sim P_{\theta}(X|Z)=Bernoulli(\p_t), \label{eq:recon}
% \end{align}
\begin{align}
\z_t&\sim Q_{\phi}(Z|X)=\NM(\muu_t,\Si_t), \;\;
\widetilde{\z}_{t+1}\sim \widetilde{Q}_{\psi}(\widetilde{Z}|Z,\u)=\NM(\A_t\muu_t+\B_t\u_t+\oo_t,\C_t), \;\;
\widetilde{\x}_t,\widetilde{\x}_{t+1}\sim P_{\theta}(X|Z)=Bern(\p_t), \label{eq:e2c_all}
\end{align}
where the last equation is the reconstruction model to be discussed later, $\C_t=\A_t\Si_t\A_t^T+\H_t$, and $\H_t$ is the covariance matrix of the estimated system noise ($\Ome_t\sim\NM(\0,\H_t)$). The key here is to learn $\A_t$, $\B_t$ and $\oo_t$, which are parameterized as follows:
\begin{align*}
\vec(\A_t)=\W_Ah_{\psi}^{\text{trans}}(\z_t)+\b_A, \;\;\;\;
\vec(\B_t)=\W_Bh_{\psi}^{\text{trans}}(\z_t)+\b_B, \;\;\;\;
\oo_t =\W_o h_{\psi}^{\text{trans}}(\z_t)+\b_o,
\end{align*}
where $h_{\psi}^{\text{trans}}(\z_t)$ is the output of the transition network.

\textbf{Reconstruction Model}:
As mentioned in the last part of Equation (\ref{eq:e2c_all}), the posterior distribution $P_{\theta}(X|Z)$ reconstructs the raw images $\x_t$ from the latent states $\z_t$. The parameters for the Bernoulli distribution $\p_t=\W_{p}h_{\theta}^{\text{dec}}(\z_t)+\b_{p}$ where $h_{\theta}^{\text{dec}}(\z_t)$ is the output of a third network, called the decoding network or the reconstruction network. Putting it all together, Equation (\ref{eq:e2c_all}) shows the generative process of the full model.

\subsubsection{Learning Using Stochastic Gradient Variational Bayes}
With $\mathcal{D}=\{(\x_1,\u_1,\x_2),\dots,(\x_{T-1},\u_{T-1},\x_T)\}$ as the training set, the loss function is as follows:
\begin{align*}
\mathcal{L}=\sum\limits_{(\x_t,\u_t,\x_{t+1})\in\mathcal{D}}\mathcal{L}^{\text{bound}}(\x_t,\u_t,\x_{t+1})
+\lambda~\mbox{KL}(\widetilde{Q}_{\psi}(\widetilde{Z}|\muu_t,\u_t)\|Q_{\phi}(Z|\x_{t+1})),
\end{align*}
where the first term is the variational bound on the marginalized log-likelihood for each data point:
\begin{align*}
\mathcal{L}^{\text{bound}}(\x_t,\u_t,\x_{t+1})=\mathbb{E}_{\substack{\z_t\sim Q_{\phi}\\ \widetilde{\z}_{t+1}\sim\widetilde{Q}_{\psi}}}
(-\log P_{\theta}(\x_t|\z_t)
-\log P_{\theta}(\x_{t+1}|\widetilde{\z}_{t+1}))+\mbox{KL}(Q_{\phi}\|P(Z)),
\end{align*}
where $P(Z)$ is the prior distribution for $Z$. With the equations above, stochastic gradient variational Bayes can be used to learn the parameters.

According to the generative process in Equation (\ref{eq:e2c_all}) and the definition in Section \ref{sec:general}, the perception variables $\Om_p=\{h_{\phi}^\text{enc}(\cdot),\W_p^+,\x_t,\muu_t,\si_t,\p_t,h_{\theta}^\text{dec}(\cdot)\}$, where $\W_p^+$ is shorthand for $\{\W_{\mu},\b_{\mu},\W_{\sigma},\b_{\sigma},\W_{p},\b_{p}\}$. The hinge variables $\Om_h=\{\z_t,\z_{t+1}\}$ and the task variables $\Om_t=\{\A_t,\B_t,\oo_t,\u_t,\C_t,\Ome_t,\W_t^+,h_{\psi}^\text{trans}(\cdot)\}$, where $\W_t^+$ is shorthand for $\{\W_A,\b_A,\W_B,\b_B,\W_o,\b_o\}$.

\subsubsection{Discussion}
The example model above demonstrates BDL's capability of learning representations that satisfy domain-specific requirements. In the case of \emph{control}, we are interested in learning representations that can capture semantic information from raw input and preserve local linearity in the space of system states. 

To achieve this goal, the BDL-based model consists of two components, a perception component to \emph{see} the live video and a control (task-specific) component to \emph{infer} the states of the dynamical system. Inference of the system is based on the mapped states and the confidence of mapping from the perception component, and in turn, the control signals sent by the control component would affect the live video received by the perception component. Only when the two components work interactively within a unified probabilistic framework can the model reach its full potential and achieve the best control performance.

Note that the BDL-based control model discussed above uses a different \emph{information exchange} mechanism from that in Section \ref{sec:recsys} and Section \ref{sec:topic_models}: it follows the VAE mechanism and uses neural networks to \emph{separately} parameterize the mean and covariance of hinge variables (e.g., in the encoding model, the hinge variable $\z_t\sim\NM(\muu_t,\mbox{diag}(\si_t^2))$, where $\muu_t$ and $\si_t$ are perception variables parameterized as in Equation (\ref{eq:encode_all})), which is more flexible (with more free parameters) than models like CDL and CDR in Section \ref{sec:recsys}, where Gaussian distributions with fixed variance are also used. Note that this BDL-based control model is an LV model as shown in Table \ref{table:summary}, and since the covariance is assumed to be diagonal, the model still meets the independence requirement in Section \ref{sec:general}.

\subsection{Bayesian Deep Learning for Other Applications}
BDL has found wide applications such as recommender systems, topic models, and control in supervised learning, unsupervised learning, and representation learning in general. In this section, we briefly discuss a few more applications that could benefit from BDL.

% Besides, xxxx, in this , briefly discuss other applications including computer vision, natural language processing, speech recognition, health care, and link prediction.

\subsubsection{Link Prediction}
Link prediction has long been a core problem in network analysis and is recently attracting more interest with the new advancements brought by BDL and deep neural networks in general. \cite{RDL} proposed the first BDL-based model, dubbed relational deep learning (RDL), for link prediction. Graphite~\cite{Graphite} extends RDL using a perception component based on graph convolutional networks (GCN)~\cite{GCN}. \cite{SBGNN} combines the classic stochastic blockmodel~\cite{SBM} (as a task-specific component) and GCN-based perception component to jointly model latent community structures and link generation in a graph with reported state-of-the-art performance in link prediction.  
%latent structures such as communities, conditional dependencies between node features and edges
% \cite{Graphite} \cite{RDL} \cite{SBGNN} \cite{SBM} \cite{GCN} deep Generative Latent FeatureRelational Model),

\subsubsection{Natural Language Processing}
Besides topic modeling as discussed in Section~\ref{sec:topic_models}, BDL is also useful for natural language processing in general. For example, \cite{S2BS} and \cite{QuaSE} build on top of the BDL principles to define a language revision process. These models typically involve RNN-based perception components and relatively simple task-specific components linking the input and output sequences. 

\subsubsection{Computer Vision}
BDL is particularly powerful for computer vision in the unsupervised learning setting. This is because in the BDL framework, one can clearly define a generative process of how objects in a scene are generated from various factors such as counts, positions, and the content~\cite{AIR}. The perception component, usually taking the form of a probabilistic neural network, can focus on modeling the raw images' visual features, while the task-specific component handles the conditional dependencies among objects' various attributes in the images. One notable work in this direction is \emph{Attend, Infer, Repeat} (AIR)~\cite{AIR}, where the task-specific component involves latent variables on each object' position, scale, appearance, and presence (which is related to counting of objects). Following AIR, variants such as Fast AIR~\cite{FastAIR} and Sequential AIR~\cite{SQAIR} are proposed to improve its computational efficiency and performance. Besides unsupervised learning, BDL can also be useful for supervised learning tasks such as action recognition in videos~\cite{ATF}, where conditional dependencies among different actions are modeled using a task-specific component.

\subsubsection{Speech}
In the field of speech recognition and synthesis, researchers have also been adopting the BDL framework to improve both accuracy and interpretability. For example, factorized hierarchical VAE~\cite{FHVAE,SFHVAE} composes VAE with a factorized latent variable model (represented as a PGM) to learn different latent factors in speech data following an unsupervised setting. Similarly, Gaussian mixture VAE~\cite{GMVAE} uses a Gaussian mixture model as the task-specific component to achieve controllable speech synthesis from text. In terms of speech recognition, recurrent Poisson process units (RPPU)~\cite{RPPU} instead adopt a different form of task-specific component; they use a stochastic process (i.e., a Poisson process) as the task-specific component to model boundaries between phonemes and successfully achieve a significantly lower word error rate (WER) for speech recognition. Similarly, deep graph random process (DGP)~\cite{DGP} as another stochastic process operates on graphs to model the relational structure among utterances, further improving performance in speech recognition.
% Audio data is similar to image and video data in that they are all 

\subsubsection{Time Series Forecasting}
Time series forecasting is a long-standing core problem in economics, statistics, and machine learning~\cite{harvey1990forecasting}. It has wide applications across multiple areas. For example, accurate forecasts of regional energy consumption can provide valuable guidance to optimize energy generation and allocation. In e-commerce, retails rely on demand forecasts to decide when and where to replenish their supplies, thereby avoiding items going out of stock and guaranteeing fastest deliveries for customers. During a pandemic such as COVID-19, it is crucial to obtain reasonable forecasts on hospital workload and medical supply demand in order to best allocate resources across the country. Needless to say, an ideal forecasting model requires both efficient processing of high-dimensional data and sophisticated modeling of different random variables, either observed or latent. BDL-based forecasting models~\cite{DeepAR,SQF-RNN,DeepState,DeepFactor} achieve these with an RNN-based perception component and a task-specific component handling the conditional dependencies among different variables, showing substantial improvement over previous non-BDL forecasting models.

\subsubsection{Health Care}
In health-care-related applications~\cite{ravi2016deep}, it is often desirable to incorporate human knowledge into models, either to boost performance or more importantly to improve interpretability. It is also crutial to ensure models' robustness when they are used on under-represented data. BDL therefore provides a unified framework to meet all these requirements: (1) with its Bayesian nature, it can impose proper priors and perform Bayesian model averaging to improve robustness; (2) its task-specific component can naturally represents and incorporate human knowledge if necessary; (3) the model's joint training provides interpretability for its both components. For example, \cite{DPFM} proposed a deep Poisson factor model, which essentially stacks layers of Poisson factor models, to analyze electronic health records. \cite{BBFDR} built a BDL model with experiment-specific priors (knowledge) to control the false discovery rate during study analysis with applications to cancer drug screening. \cite{DeepMarkov} developed deep nonlinear state space models and demonstrated their effectiveness in processing electronic health records and performing counterfactual reasoning. Task-specific components in the BDL models above all take the form of a typical Bayesian network (as mentioned in Section~\ref{sec:bayesnet}). In contrast, \cite{BIN} proposed to use bidirectional inference networks, which are essentially a class of deep Bayesian network, as the task-specific component (as mentioned in Section~\ref{sec:bin}). This enables deep nonlinear structures in each conditional distribution of the Bayesian network and improves performance for applications such as health profiling.

%\section{Future Research}\label{sec:future}
\section{Conclusions and Future Research}\label{sec:summary}
BDL strives to combine the merits of PGM and NN by organically integrating them in a single principled probabilistic framework. In this survey, we identified such a current trend and reviewed recent work. A BDL model consists of a perception component and a task-specific component; we therefore surveyed different instantiations of both components developed over the past few years respectively and discussed different variants in detail. To learn parameters in BDL, several types of algorithms have been proposed, ranging from block coordinate descent, Bayesian conditional density filtering, and stochastic gradient thermostats to stochastic gradient variational Bayes.

BDL draws inspiration and gain popularity both from the success of PGM and from recent promising advances on deep learning. Since many real-world tasks involve both efficient perception from high-dimensional signals (e.g., images and videos) and probabilistic inference on random variables, BDL emerges as a natural choice to harness the perception ability from NN and the (conditional and causal) inference ability from PGM.
Over the past few years, BDL has found successful applications in various areas such as recommender systems, topic models, stochastic optimal control, computer vision, natural language processing, health care, etc. In the future, we can expect both more in-depth studies on existing applications and exploration on even more complex tasks. 
% an increasing number of other applications like link prediction, community detection, active learning, Bayesian reinforcement learning, and many other complex tasks that need interaction between perception and causal inference. 
Besides, recent progress on efficient BNN (as the perception component of BDL) also lays down the foundation for further improving BDL's scalability. 
% with the advances of efficient \emph{Bayesian neural networks} (BNN), BDL with BNN as an important component is expected to be more and more scalable.

\bibliography{survey,hao,bnn,bdl,nn}
\bibliographystyle{plain}

\end{document}